%% file: main.tex
\documentclass{article}

\usepackage{arxiv}
\pdfoutput=1

\usepackage[utf8]{inputenc} 
\usepackage[T1]{fontenc}    
\usepackage{url}            
\usepackage{booktabs}       
\usepackage{amsfonts}       
\usepackage{nicefrac}       
\usepackage{microtype}      
\usepackage{lipsum}
\usepackage{enumitem}
\usepackage{natbib}
\usepackage{url}            
\usepackage{graphicx}
\usepackage{subfigure}
\usepackage[skins]{tcolorbox}
\usepackage{hyperref}
\usepackage[T1]{fontenc}
\usepackage{epigraph}
\usepackage[export]{adjustbox}
\usepackage{amssymb,mathtools}
\usepackage{multirow}
\usepackage{enumitem}
\usepackage{amsmath}
\usepackage{hyperref}

\hypersetup{
    colorlinks=true,
    linkcolor=blue,
    filecolor=magenta,      
    urlcolor=blue,
    citecolor=blue
}

\makeatletter
\newcommand{\printfnsymbol}[1]{%
  \textsuperscript{\@fnsymbol{#1}}%
}
\makeatother

\input{math_commands.tex}

\usepackage{eqparbox}

\newcommand{\lefteqbox}[2]{%
  \eqparbox[t]{#1}{$\displaystyle#2$\hfil}%
}
\usepackage{caption} 
\usepackage{algorithm}
\usepackage[noend]{algorithmic}
\renewcommand{\algorithmiccomment}[1]{\bgroup\hfill $\triangleright$ ~#1\egroup}

\newcommand{\todocite}[1]{\textcolor{brown}{[ref]}}

\newcommand{\methodname}{\emph{SimpleBits}~}
\newcommand{\bpd}{$\mathrm{bpd}$~}

\definecolor{darkgreen}{rgb}{0.0, 0.28, 0.2}

\newcommand{\titl}{When less is more: Simplifying inputs aids neural network understanding}
\title{\titl}

\newcommand{\algcomment}[1]{\textcolor{gray}{\textit{#1}}}

\author{Robin Tibor Schirrmeister  \\
University Medical Center Freiburg\\
ML Collective \\
\texttt{robin.schirrmeister@uniklinik-freiburg.de} \\
\And
Tonio Ball \\
University Medical Center Freiburg\\
\texttt{tonio.ball@uniklinik-freiburg.de} \\
\And
Sara Hooker \\
Google Brain \\
ML Collective \\
\texttt{shooker@google.com} \\
\And
Rosanne Liu \\
Google Brain \\
ML Collective \\
\texttt{rosanneliu@google.com} \\
}

\begin{document}

\maketitle

\begin{abstract}

How do neural network image classifiers respond to simpler and simpler inputs? And what do such responses reveal about the learning process? To answer these questions, we need a clear measure of input simplicity (or inversely, complexity), an optimization objective that correlates with simplification, and a framework to incorporate such objective into training and inference. Lastly we need a variety of testbeds to experiment and evaluate the impact of such simplification on learning. 
In this work, we measure simplicity with the encoding bit size given by a pretrained generative model, and minimize the bit size to simplify inputs in training and inference. We investigate the effect of such simplification in several scenarios: conventional training, dataset condensation and post-hoc explanations. In all settings, inputs are simplified along with the original classification task, and we investigate the trade-off between input simplicity and task performance. For images with injected distractors, such simplification naturally removes superfluous information. For dataset condensation, we find that inputs can be simplified with almost no accuracy degradation. When used in post-hoc explanation, our learning-based simplification approach offers a valuable new tool to explore the basis of network decisions.

\end{abstract}



\section{Introduction}

A better understanding of the information deep neural networks use to learn can lead to new scientific discoveries \citep{raghu2020survey}, highlight differences between human and model behaviors \citep{makino2020differences} and serve as powerful auditing tools \citep{geirhos2020shortcut,dsouza2021tale,bastings2021will,agarwal2021estimating}.

Removing information from the input deliberately is one way to illuminate what information content is relevant for learning. For example, occluding specific regions, or removing certain frequency ranges from the input gives insight into which input regions and frequency ranges are relevant for the network's prediction \citep{DBLP:conf/iclr/ZintgrafCAW17,makino2020differences,DBLP:journals/corr/abs-2107-10356}. These ablation techniques use simple heuristics such as random removal \citep{NEURIPS2019_fe4b8556,madsen2021evaluating}, or exploit domain knowledge about interpretable aspects of the input to create simpler versions of the input on which the network's prediction is analyzed~\citep{banerjee2021reading}.

What if, instead of using heuristics, one \emph{learns} to synthesize simpler inputs that retrain prediction-relevant information? This way, we could gain intuition into the model behavior without relying on prior or domain knowledge about what input content may be relevant for the network's learning target. To achieve this, one needs to define the precise meaning of “simplifying an input”, including direct metrics for such simplification and retention of task-relevant information.

In this work, we propose  \emph{SimpleBits}, an information-reduction method that learns to synthesize simplified inputs which contain less information while remaining informative for the task. To measure simplicity, we use a finding initially reported as a problem for density-based anomaly detection---that generative image models tend to assign higher probability densities and hence lower bits to visually simpler inputs \citep{DBLP:conf/nips/KirichenkoIW20,DBLP:conf/nips/SchirrmeisterZB20}. However, we use this to our advantage and minimize the encoding bit size given by a generative network trained on a general image distribution to simplify inputs. At the same time, we optimize for the original task so that the simplified inputs still preserve task-relevant information.

\begin{figure*}[ht!]
    \centering
    \includegraphics[width=0.85\linewidth]{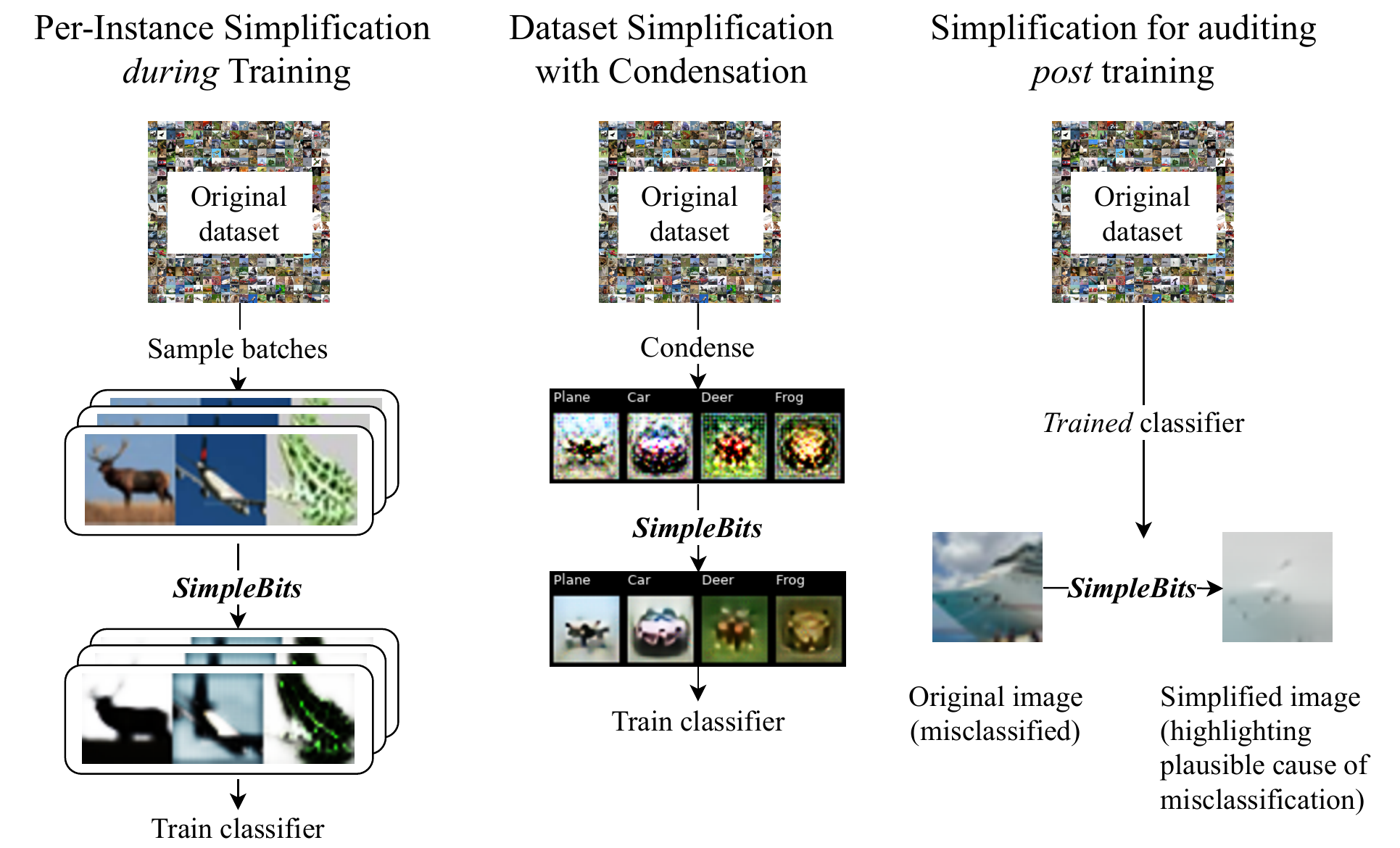}
    \caption{We apply \methodname to a variety of tasks to aid neural network understanding. \textbf{Left:} \methodname acts as a per-instance simplifier during training, where each image is simplified but the total number of images remain the same.    \textbf{Center:} \methodname is combined with data condensation techniques, to reduce both the size of the dataset and the complexity of individual images in the condensed dataset. 
    \textbf{Right:} \methodname is used as a post-hoc analysis tool of any trained classifier, illuminating features of images that are crucial to the network's decision. }
    \label{fig:overview}
\end{figure*}

We apply \methodname both in a \emph{per-instance} setting, where each image is processed to be a simplified version of itself, and the size of training set remains unchanged, as well as in a \emph{condensation} setting, where the dataset is compressed to only a few samples per class, with the condensed samples simplified at the same time. Applied during training, \methodname can be used to investigate the trade-off between the information content and the task performance. Post training, \methodname can act as an analysis tool to understand what information a trained model uses for its decision making. \figref{fig:overview} summarizes the tasks covered in this paper.

Our evaluation provides the following insights in the investigated scenarios:
\begin{enumerate}
    \item \textbf{Per-instance simplification during training.} \methodname successfully removes superfluous information for tasks with injected distractors. On natural image datasets, \methodname highlights plausible task-relevant information (shape, color, texture). Increasing simplification leads to accuracy decreases, as expected, and we report the trade-off between input simplification and task performance for different datasets and problems.
    \item \textbf{Dataset simplification with condensation.} We evaluate \methodname applied to a condensation setting where the training data are converted into a much smaller set of synthetic images. \methodname further simplifies these images to drastically reduce the encoding size without substantial task performance decrease. On a chest radiograph dataset  \citep{johnson2019mimic,johnson2019mimicjpg}, \methodname can uncover known radiologic features for pleural effusion and gender. 
    \item \textbf{Simplification for auditing post-training.} Given a trained classifier on regular datasets, we explore the use of \methodname as an interpretability tool for auditing processes. Our exploration of \methodname guided audits suggests it can provide intuition into model behavior on individual examples, including finding features that may contribute to misclassifications. 
    
\end{enumerate}


\begin{figure}[ht!]
    \centering
    \includegraphics[width=\linewidth]{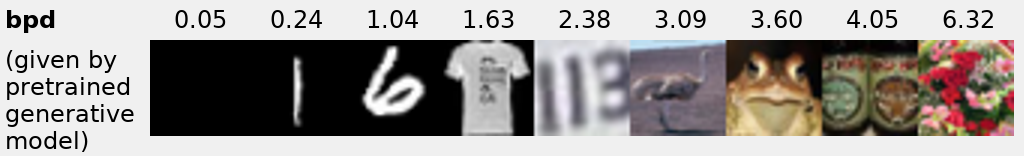}
    \caption{Visualization of the bits-per-dimension ($\mathrm{bpd}$) measure for image complexity, sorted from low to high. Image samples are taken from MNIST, Fashion-MNIST, CIFAR10 and CIFAR100, in addition to a completely black image sample. \bpd is calculated from the density produced by a Glow~\citep{NIPS2018_8224} model pretrained on 80 Million Tiny Images. Switching to other types of generative models including PixelCNN and diffusion models trained on other datasets produces consistent observations; see \figref{suppfig:bpds-other-models} in Supplementary Information for more details.
   }
    \label{fig:bits-complexity}
\end{figure}



\section{Measuring and Reducing Instance Complexity}\label{subsec:measure_instance_complexity}

How to define simplicity? We use the fact that generative image models tend to assign lower encoding bit sizes to visually simpler inputs \citep{DBLP:conf/nips/KirichenkoIW20,DBLP:conf/nips/SchirrmeisterZB20}. Concretely, the complexity of an image $\bm{x}$ can be quantified as the negative log probability mass given by a pretrained generative model with tractable likelihood, $G$, i.e. $-\log p_G(\bm{x})$. $-\log p_G(\bm{x})$ can be interpreted as the image encoding size in bits per dimension (bpd) via Shannon’s theorem \citep{Shannon1948}:  $\textrm{bpd}(\bm{x})=-\log_2 p_G(\bm{x})/d$ where $d$ is the dimension of the flattened $\bm{x}$. 

The simplification loss for an input $\bm{x}$, given a pre-trained generative model $G$, is as follows:
\begin{align}
\lefteqbox{A}{L_{\mathrm{sim}}}(\bm{x})&=-\log p_G(\bm{x}) \label{eq:simplification-loss}
\end{align}

\figref{fig:bits-complexity} visualizes images and their corresponding bits-per-dimension ($\mathrm{bpd}$) values given by a Glow network \citep{NIPS2018_8224} trained on 80 Million Tiny Images \citep{80mtiny} (see Supplementary Section~\ref{suppsec:bpds-other-models-network} for other models). This is the generative network used across all our experiments. A visual inspection of \figref{fig:bits-complexity} suggests that lower \bpd corresponds with simpler inputs, as also noted in prior work \citep{DBLP:conf/iclr/SerraAGSNL20}. The goal of our approach, \emph{SimpleBits}, is to minimize \bpd of input images while preserving task-relevant information. 

We now explore how \methodname affects network behavior in a variety of scenarios. In each scenario, we explain the method to optimize $L_{\mathrm{sim}}$ and the retention of task-relevant information, the experimental setup, and results. All code is at \url{https://tinyurl.com/simple-bits}.

\section{Per-instance simplification during training}\label{subsec:instance-simplification}

When plugged into the training of a classifier $f$, \methodname simplifies each image such that $f$ can still learn the original classification task from the simplified batches. We apply backpropagation through training steps: given a batch of input images $\bm{X}_{\mathrm{orig}}$, before updating the classifier $f$, an image-to-image $\mathrm{simplifier}$ network generates a corresponding batch of images $\bm{X}_{\mathrm{sim}}$ such that: \textbf{(a)} images in $\bm{X}_{\mathrm{sim}}$ have low \bpd as measured per $L_\mathrm{sim}$ in ~\Eqref{eq:simplification-loss}, and \textbf{(b)} training on the simplified images leads to a reduction of the classification loss on the original images.

We optimize \textbf{(b)} by unrolling one training step of the classifier. So for a single batch $\bm{X}_\mathrm{orig}, \bm{y}$, we first compute an updated classifier $f'$ as follows:

\begin{align}
\lefteqbox{A}{\bm{X}_\mathrm{sim}} &=\mathrm{simplifier}(\bm{X}_\mathrm{orig})\\
\lefteqbox{A}{f'}&=\mathrm{train\_step}(f, l(f(\bm{X}_{\mathrm{sim}}), \bm{y}))
\label{eq:classification-loss}
\end{align}

where $\mathrm{simplifier}$ is an image-to-image network, $l$ is the classification loss function, i.e., the cross-entropy loss, and $f'$ is the classifier $f$ after one optimization step using the classification loss $l(f(\bm{X}_{\mathrm{sim}}), \bm{y})$ on the simplified data.

To train the $\mathrm{simplifier}$ network, we  optimize for both \textbf{(a)} (\Eqref{eq:simplification-loss}) and for the classification loss on the batch of original images with the updated classifier after the unrolled training step $l(f'(\bm{X}_\mathrm{orig}), \bm{y})$ using backpropagation through training \citep{pmlr-v37-maclaurin15,DBLP:conf/icml/FinnAL17}. Note that one could not instead directly optimize performance on the simplified images as the simplifier could then collapse all simplified images of one class to one representative example.
\begin{figure*}[ht!]
    \centering
    \includegraphics[width=0.9\linewidth]{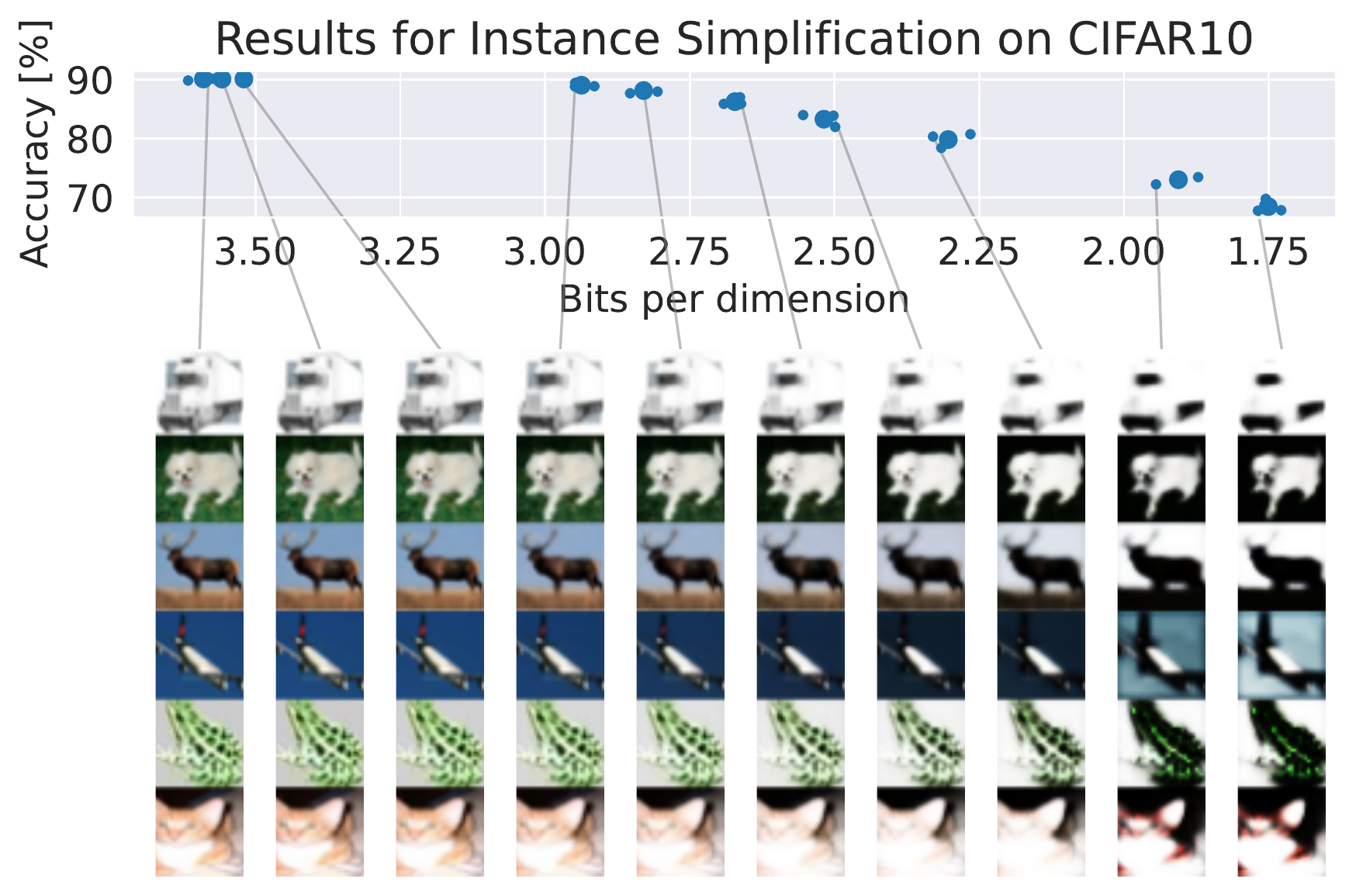}      
    \caption{Selection of \methodname applied to CIFAR10 training images. Simplified images from settings with varying simplification loss weight $\lambda_\textrm{sim}$. We observe that at lower bits per dimension color is retained only for some images such as green color for frog or blue sky for the plane. In this low bit regime, texture remains discernable for the cat and frog images.  More simplified images are in \secref{suppsec:images-during-training}.}
    \label{fig:results-instance-cifar10}
\end{figure*}

Adding the classification losses on the simplified data both before and after the unrolled training step ensures the training is not influenced by predictions on the simplified data that are very different from the classification target and we found that to improve training stability, leading to: 
\begin{align}
\lefteqbox{A}{L_{\mathrm{cls}}}&=l(f'(\bm{X}_\mathrm{orig}), \bm{y}) + l(f(\bm{X}_\mathrm{sim}),\bm{y}) + l(f'(\bm{X}_\mathrm{sim}), \bm{y})
\end{align}

Further details to stabilize the training are described in Supplementary \ref{suppsec:instance-optimization}. The total loss for the $\mathrm{simplifier}$ is
\begin{align}
L&=\lambda_\textrm{sim} \cdot L_{\mathrm{sim}} + L_{cls},
\end{align}
where $\lambda_\textrm{sim}$ is a hyperparameter for the trade-off between simplification and task performance. In subsequent experiments, we vary $\lambda_\textrm{sim}$ to control the extent of simplification. 


\textbf{Implementation} Our training architecture is a normalizer-free classifier architecture to avoid interpretation difficulties that may arise from normalization layers, such as image values being downscaled by the $\mathrm{simplifier}$ and then renormalized again. We use Wide Residual Networks as described by \citet{DBLP:conf/icml/BrockDSS21} for the classifier. The normalizer-free architecture reaches 94.0\% on CIFAR10 in our experiments, however we opt for a smaller variant for faster experiments that reaches 91.2\%; additional details are included in the Supplementary Section \ref{suppsec:classifier-network}. For the $\mathrm{simplifier}$ network, we use a UNet architecture \citep{DBLP:conf/miccai/RonnebergerFB15} that we modify to be residual, more details included in Supplementary \secref{suppsec:simplifier-network}.
For both $\mathrm{simplifier}$ and classifier networks, we use AdamW \citep{DBLP:conf/iclr/LoshchilovH19} with $\mathrm{lr}=5\cdot10^{-4}$ and $\mathrm{weight\_decay}=10^{-5}$.

\subsection{\methodname Removes Injected Distractors}\label{subsec:redundant-features}

\begin{figure*}[ht!]
    \centering
    \includegraphics[width=\linewidth]{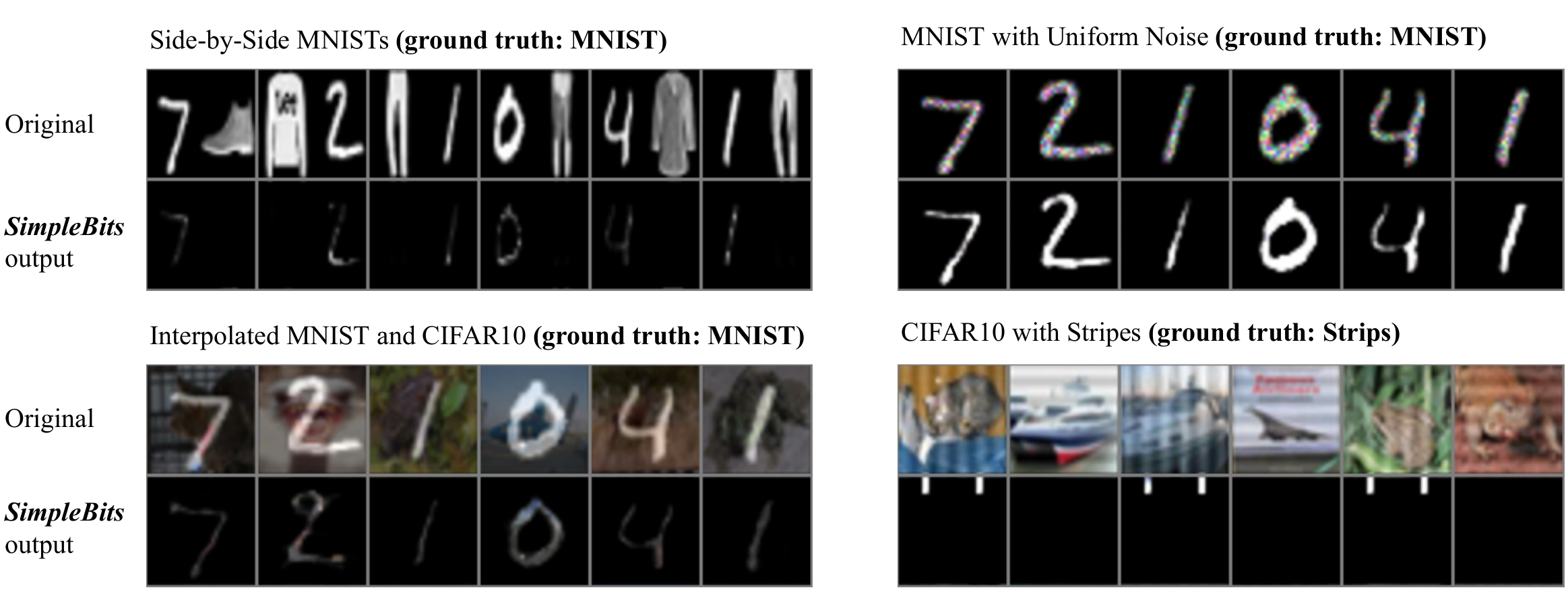}
    \caption{Evaluation of \methodname as a distractor removal on four composite datasets. Shown are the composite original images and the corresponding simplified images produced by \methodname trained alongside the classifier. \methodname is able to almost entirely remove task-irrelevant image parts, namely FashionMNIST (\textbf{top left}), random noise (\textbf{top right}), CIFAR10 (\textbf{bottom left} as well as \textbf{bottom right}).} 
    \label{fig:results-instance-synthetic}
\end{figure*}

We first evaluate whether our per-instance simplification during training successfully removes superfluous information for tasks with injected distractors. To that end, we construct datasets to contain both useful (ground truth) and redundant (distractor) information for task learning. We create four composite datasets derived from three conventional datasets: MNIST \citep{lecun-mnisthandwrittendigit-2010}, FashionMNIST~ \citep{xiao2017/online} and CIFAR10 \citep{Cifar10_Krizhevsky09learningmultiple}. Sample images, both constructed (input to the whole model) and simplified (output of $\mathrm{simplifier}$ and input to classifier), are shown in~\figref{fig:results-instance-synthetic}.

\textbf{Side-by-Side MNIST} constructs each image by horizontally concatenating one image from Fashion-MNIST and one from MNIST. Each image is rescaled to 16x32, so the concatenated image size remains 32x32; the order of concatenation is random. The ground truth target is MNIST labels, and therefore FashionMNIST is an irrelevant distractor for the classification task. As seen in~\figref{fig:results-instance-synthetic}, the $\mathrm{simplifier}$ effectively removes the clothes side of the image. 

\textbf{MNIST with Uniform Noise} adds uniform noise to the MNIST digits, preserving the MNIST digit as the classification target. Hence the noise is the distractor and is expected to be removed. And indeed the noise is no longer visible in the simplified outputs shown in~\figref{fig:results-instance-synthetic}.

\textbf{Interpolated MNIST and CIFAR10} is constructed by interpolating between MNIST and CIFAR10 images. MNIST digits are the classification target. The expectation is that the simplified images should no longer contain any of the CIFAR10 image information. The result shows that most of the CIFAR10 background is removed, leaving only slight traces of colors.

\textbf{CIFAR10 with Stripes} overlays either horizontal or vertical stripes onto CIFAR10 images, with the binary classification label 0 for horizontal and 1 for vertical stripes. With this dataset we observe the most drastic and effective information removal, where only the tip of vertical strips is retained, which by itself is sufficient to solve this binary classification task.

\begin{figure}
    \centering
    \includegraphics[width=\linewidth]{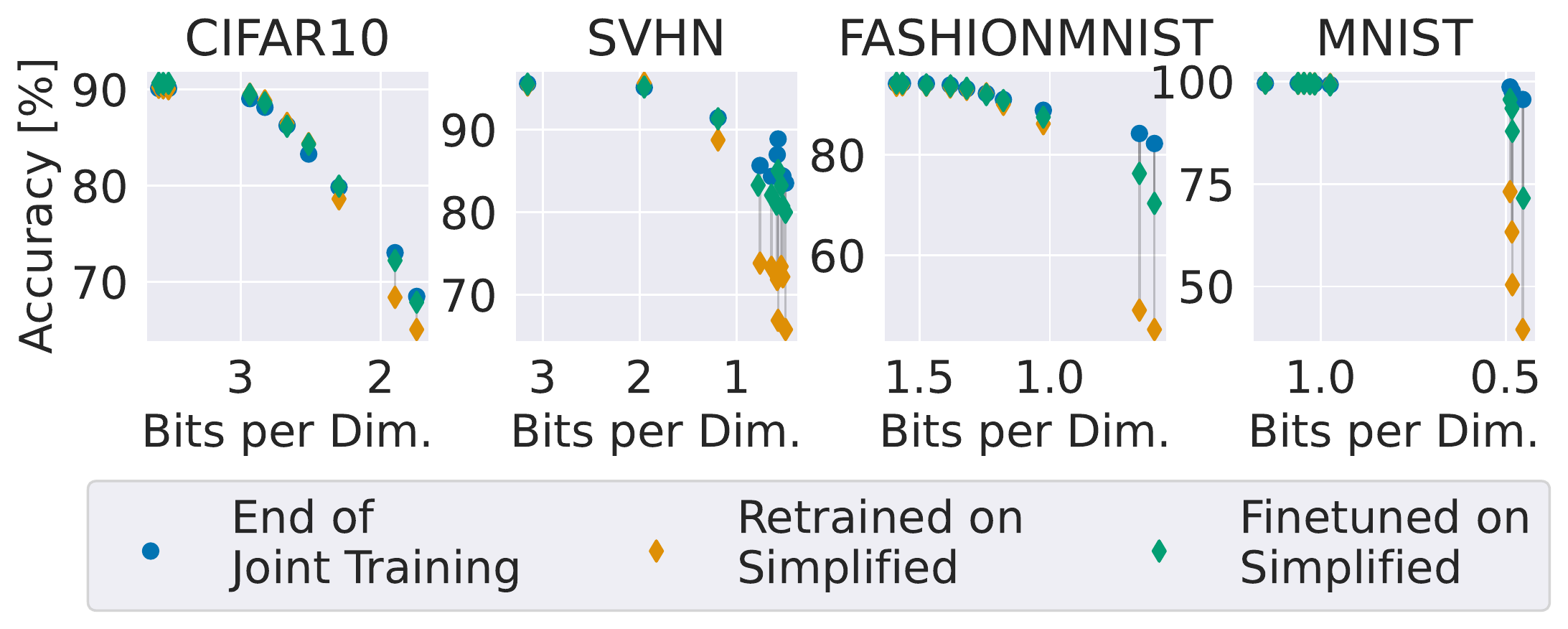}
     \vspace{-0.5cm}
    \caption{Results for training image simplifications on real datasets. Dots show results for training with different loss weights for the simplification loss. Images with less bits per dimension lead to reduced accuracies, and such reduction is more pronounced for complex datasets like CIFAR10.}
    \label{fig:results-instance-real-datasets}
    \vspace{-10pt}
\end{figure}

\subsection{Trade-off Curves on Conventional Datasets}\label{subsec:trade-off-curves}

We verified that \methodname is able to discern information relevance in inputs, and effectively removes redundant content to better serve the classification task. In real-world datasets, however, the type, as well as the amount of information redundancy in inputs is often unclear. 

With the same framework, by varying the simplification weight $\lambda_\textrm{sim}$ we can study the trade-off between task performance and level of simplification to better understand a given dataset. On one end, with $\lambda_\textrm{sim}=0$ it should faithfully restores to conventional training. On the other end, when $\lambda_\textrm{sim}$ is sufficiently high (resulting in an all-black input image), the training is expected to fail.

We experimented with MNIST, Fashion-MNIST, SVHN \citep{Netzer_SVHN} and CIFAR10, producing a trade-off curve for each by setting the strength of the simplification loss to various values during training, including $\lambda_\textrm{sim}=0$. For each setting, we report the classification accuracy as a result of joint training of simplification and classification, retraining from scratch with only simplified images, and finetuning after joint training with simplified images, respectively, as shown in \figref{fig:results-instance-real-datasets}.

As expected, higher strengths of simplification lead to decreased task performance. Interestingly, such decay is observed to be more pronounced for more complex datasets such as CIFAR10. This suggests either the presence of a relatively small amount of information redundancy,
or that naturally occurring noise in data helps with generalization, analogous to how injected noise from data augmentation helps learning features invariant to translations and viewpoints, even though the augmentation itself does not contain additional discriminative information.


We run three baselines to validate the efficacy of our simplification framework, as described in Supplementary \secref{suppsec:simplifier-mse-baseline} and shown in \figref{suppfig:simplifier-mse-baseline}. \secref{suppsec:png-storage-space} shows that our simplification has practical savings in image storage.
More analyses including training curves and potential spurious features revealed by \methodname can be seen in Supplementary \secref{suppsec:retrain-learning-curves} and  \ref{suppsec:spurious-features}, respectively.



\section{Dataset simplification with Condensation}\label{subsec:dataset-condensation}

Now we investigate how \methodname affects training on a small synthetic condensed dataset. Multiple methods have been developed for dataset condensation \citep{DBLP:conf/iclr/ZhaoMB21, DBLP:conf/icml/ZhaoB21,DBLP:journals/corr/abs-1811-10959,pmlr-v37-maclaurin15}, via backpropagation through training \citep{DBLP:journals/corr/abs-1811-10959,pmlr-v37-maclaurin15},  gradient matching~\citep{DBLP:conf/icml/ZhaoB21}, or kernel based meta-learning \citep{DBLP:journals/corr/abs-2107-13034}. Due to its small size, one can visualize the full condensed dataset to understand what information is preserved for learning.  Our aim here is to combine \methodname with dataset condensation to see if we could obtain a both smaller and  simpler training dataset than the original.

In this setting, we jointly condense our training dataset to a smaller number of synthetic training inputs and simplify the synthetic inputs according to our simplification loss (Eq. \ref{eq:simplification-loss}). 
Concretely, we add the simplification loss $L_{\mathrm{sim}}$ to the gradient matching loss proposed by \citet{DBLP:conf/icml/ZhaoB21}. The gradient matching loss computes the layerwise cosine distance between the gradient of the classification loss wrt. to the classifier parameters $\theta$ produced by a batch of original images $\bm{X}_\mathrm{orig}$ and a batch of synthetic images $\bm{X}_\mathrm{syn}$:

\begin{align}
\begin{split}
    &L_\textrm{match}(\bm{X}_\mathrm{orig},\bm{X}_\mathrm{syn})  =\\ &D(\nabla_{\theta}l(f(\bm{X}_\mathrm{orig}),\bm{y}),  \nabla_{\theta}l(f(\bm{X}_\mathrm{syn}),\bm{y})). 
\label{eq.optgrad}
\end{split}
\end{align}

where $D$ is the layerwise cosine distance. The matching loss is computed separately per class.

Overall, with our simplification loss, we get:

\begin{align}
\begin{split}
    L_{\mathrm{syn}}( \bm{X}_\mathrm{orig},\bm{X}_\mathrm{syn}) = & \  L_{\mathrm{match}}(\bm{X}_\mathrm{orig},\bm{X}_\mathrm{syn})\\ 
    &  + 
    \sum_{\bm{x}_\mathrm{syn} \in \bm{X}_{\mathrm{syn}}}-\log{p_G(\bm{x}_{\mathrm{syn}})}
\end{split}
\end{align}


We perform dataset condensation on MNIST, Fashion-MNIST, SVHN and CIFAR10 with varying $\lambda_\textrm{sim}$ for the simplification loss. We also apply dataset condensation to the chest radiograph dataset MIMIC-CXR-JPG \citep{johnson2019mimic,johnson2019mimicjpg} for predicting pleural effusion and gender. We use the networks from \citep{DBLP:conf/iclr/ZhaoMB21}, but use Adam \citep{DBLP:journals/corr/KingmaB14} for optimization.

\subsection{\methodname retains condensation performance while greatly simplifying data}

\begin{figure*}
    \centering
    \includegraphics[width=0.8\linewidth]{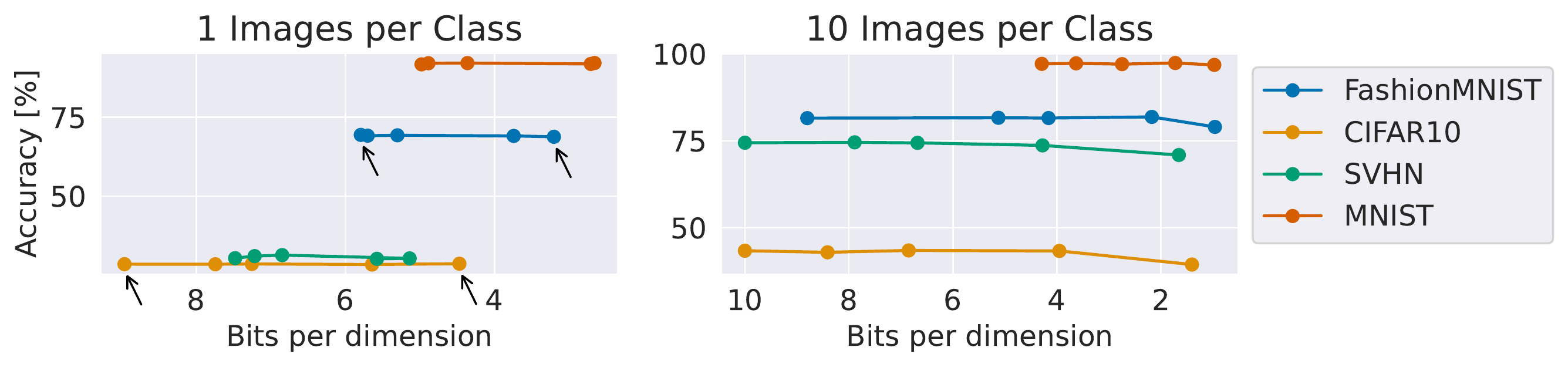}
    \centering
    \includegraphics[width=0.9\linewidth]{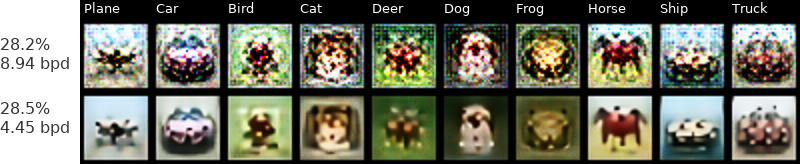}
    \includegraphics[width=0.9\linewidth]{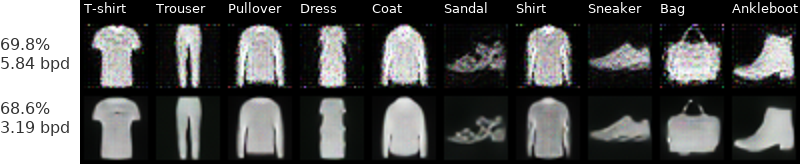}
    \caption{
 Dataset condensation accuracies (when retraining with the condensed dataset) vs. data simplicity. \textbf{Top:} Each dot represents a data condensation experiment run with a particular weight for the simplification loss, which results in more or less complex datasets. Accuracies can be retained even with substantially reduced bits per dimension. In the 1-image-per-class case (\textbf{top left figure}), arrows highlight the settings that are visualized in the bottom figure. 
 \textbf{Bottom:} Condensed datasets with varying simplification loss weight. Each row represents the whole condensed dataset (1 image per class), with high (top row) or low (bottom row) bits per dimension. Lower bits per dimension datasets are visually simpler and smoother while retaining the accuracy.}
    \label{fig:results-condensation}
\end{figure*}

\begin{figure*}
\centering
\includegraphics[width=0.48\linewidth]{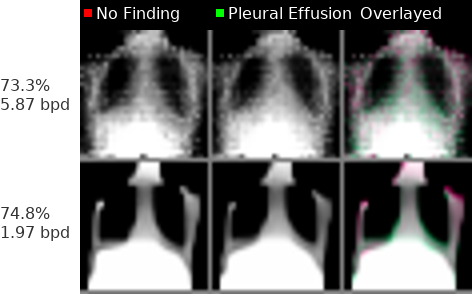}
\includegraphics[width=0.48\linewidth]{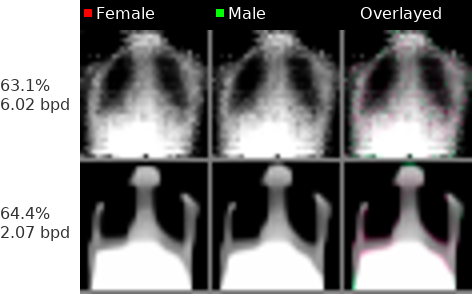}
\caption{
Condensed dataset for pleural effusion and gender prediction from chest radiographs in MIMIC-CXR. Condensed images for the classes look very similar. Color-coded mixed rightmost images reveal the differences between the classes. Green highlighted region at the lower end of the lung consistent with typical radiologic features for pleural effusion (white region indicating fluid on lungs), red highlighted around lung for gender indicate smaller lung volume for the female class.
}
\label{fig:results-mimic-cxr}
\end{figure*}

In~\figref{fig:results-condensation}, we examine the accuracy for each condensed-and-simplified dataset. We observe that for the natural image datasets, accuracies are mostly retained when decreasing the number of bits per image. Note that the setting with highest bpd is a reimplementation of \citet{DBLP:conf/iclr/ZhaoMB21} and therefore a baseline without simplification loss. We visualize examples in~\figref{fig:results-condensation} and observe that the jointly condensed and simplified images look visually smoother, indicating that higher frequency patterns visible in the original images are not needed to reach the same accuracy. These visualizations are also noticeably more smooth than the results for per-instance simplification \figref{fig:results-instance-synthetic}, which suggests that data condensation may already favor features that are less complex. Further condensed sets are in \secref{suppsec:images-condensed} and a continual learning evaluation in \ref{suppsec:continual-learning}.

\textbf{Evaluation of a medical chest radiograph dataset} We also evaluate jointly condensing and simplifying for a dataset of chest radiograph images \citep{johnson2019mimic,johnson2019mimicjpg}. This dataset has known radiologic features for the presence of pleural effusion \citep{jany2019pleural,raasch1982pleural} and difference in gender \citep{bellemare2003sex}. In \figref{fig:results-mimic-cxr}, we visualize both the condensed \textbf{(top row)} and the jointly condensed and simplified dataset \textbf{(bottom row)}. The overlayed shows that a visible difference between presence of feature. For pleural effusion, a larger white region on the bottom of the lung occurs in the simplified pathological image, while for gender, lungs appear slightly smaller for the simplified female image.

\section{Post-training  simplification}\label{subsec:post-hoc}

\begin{algorithm}
\caption{Simplification loss function after training}
\footnotesize
\label{alg:post_hoc}
\begin{algorithmic}

\STATE{\textbf{given} generative network $G$, input $\bm{x}_\mathrm{orig}$, simplified input $\bm{x}_\mathrm{sim}$, classifier $f$, parameter scaling factors $\bm{s} < 1$}
\STATE{\algcomment{Scale classifier parameters down by $\bm{s}$ to simulate "forgetting"}}
\STATE{$f_{\mathrm{scd}} \leftarrow  \text{ScaleParameters}(f,\bm{s})$} 
\STATE{\algcomment{Predict original and simplified with unscaled classifier}}
\STATE{$\bm{h}_\mathrm{orig}=f(\bm{x}_\mathrm{orig}),\bm{h}_\mathrm{sim}=f(\bm{x}_\mathrm{sim})$}
\STATE{\algcomment{Predict original and simplified with scaled classifier}}
\STATE{$\bm{h}_\mathrm{orig,scd}=f_\mathrm{scd}(\bm{x}_\mathrm{orig}),\bm{h}_\mathrm{sim,scd}=f_\mathrm{scd}(\bm{x}_\mathrm{sim})$}
\STATE{\algcomment{Compute gradients of prediction divergence wrt. scaling factors}}
\STATE{$grad_\mathrm{orig}=\nabla_{\bm{s}} \KL (\bm{h}_\mathrm{orig} \Vert \bm{h}_{\mathrm{orig,scd}}) $}
\STATE{$grad_\mathrm{sim}=\nabla_{\bm{s}}\KL (\bm{h}_\mathrm{orig} \Vert \bm{h}_{\mathrm{sim,scd}})$}
\STATE{\algcomment{Compute layerwise cosine distance between gradients}}
\STATE{$L_\mathrm{grad}=D_\mathrm{cos}(grad_\mathrm{orig},grad_\mathrm{sim})$} \label{alg:gradient-distance}
\STATE{\algcomment{Compute prediction differences of original and simplified}}
\STATE{$L_\mathrm{pred}=\KL (\bm{h} \Vert \bm{h}_{\mathrm{sim}}) + \KL (\bm{h}_{\mathrm{scd}} \Vert \bm{h}_{\mathrm{sim,scd}})$}
\STATE{\algcomment{Compute needed bits for simplified input}}
\STATE{$L_\mathrm{sim}=-\log p_G(\bm{x}_\mathrm{sim}))$}
\STATE{\textbf{return} $L_\mathrm{grad} + L_\mathrm{pred} + \lambda_\mathrm{sim} \cdot L_\mathrm{sim}$ }
\end{algorithmic}
\end{algorithm}

Can \methodname be used to interpret trained classifiers \emph{after} training?  We explore simplifying images post-training as a basis for gaining intuition about model behavior. We will do so by using \methodname to visualize some of the information that would help the classifier remember what it has learned for a specific prediction.

For synthesizing the prediction-relevant information, we simulate that the classifier forgets knowledge and then synthesize a simplified input that allows the classifier to relearn the relevant knowledge for the prediction of the original input. To simulate forgetting, we scale down all parameters of the trained classifier $f$ by multiplying them with a gating value $\phi_{\textrm{scaled},i}=\phi_{i}\cdot s, s<1$. This scaling removes information from the model by bringing the learned parameter values closer to zero, is identical to weight decay and also makes the network simpler in terms of model encoding size \citep{DBLP:conf/colt/HintonC93}. To synthesize a simplified input that helps learning to restore the parameter values that are important for a specific input, we compare gradients between the original and simplified input.

Given an input, we can compute the gradients of the KL-divergence between the rescaled network's prediction $f_{\mathrm{scd}}(x)$ and the original network's prediction $f(x)$. At each iteration, the layerwise cosine distance between these two sets of gradients (one for original datapoint and the other for the simplified) is the basis of the loss used in \emph{SimpleBits}. We compute this distance only considering the gradients that are negative for the original input. Combined with the simplification loss Eq. \ref{eq:simplification-loss}, this amounts to asking \textit{what input information is needed to recover parts of the original network to restore the original prediction?} To ensure that the network is trained towards minimizing the same prediction difference on the simplified and original data, we also add a prediction difference loss  $L_\textrm{pred}$. Further details about the implementation are included in Alg. \ref{alg:post_hoc} and Supplementary \secref{suppsec:instance-after-alg}. We use the loss computed in Alg. \ref{alg:post_hoc} to optimize the latent encoding of the simplified image with Adam.

\begin{figure*}
    \centering
    \includegraphics[width=0.9\linewidth]{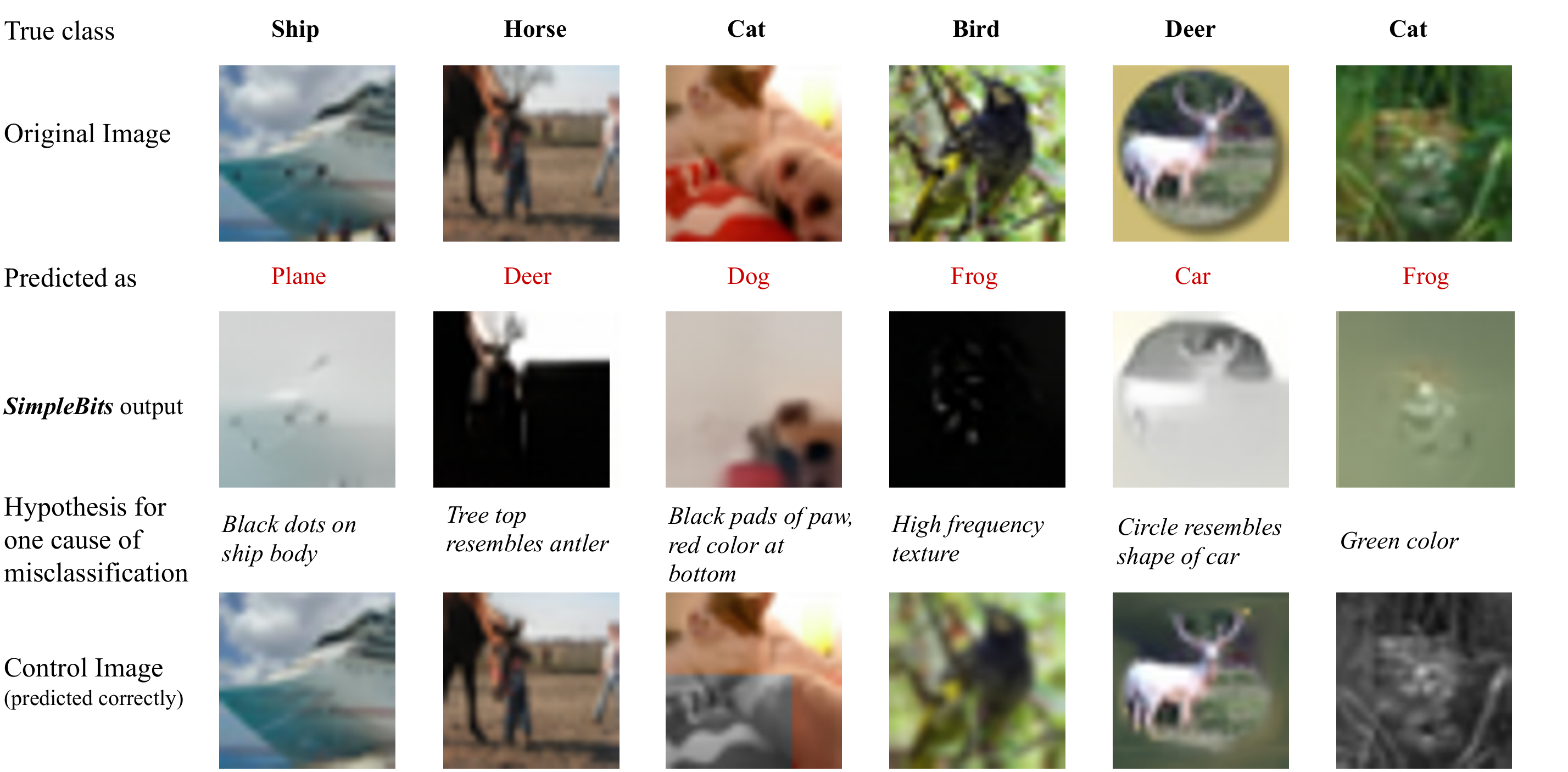}
    \caption{Post-hoc simplifications of misclassified CIFAR-10 examples. For each example, the simplified image reveals plausible causes for the misclassification. We subsequently made alterations to compensate for the cause (from left to right: removing black dots, removing tree top, removing color, removing high frequency texture, removing circle, and removing color), and are able to revert the predictions to the true class. 
    }
    \label{fig:results-post-hoc}
\end{figure*}

In \figref{fig:results-post-hoc}, we visualize both the misclassified images according to the original network $f$ and produce the corresponding simplified versions. We observe that simplified images may provide some intuition into the reason for the network's misclassification, highlighting a variety of different features for different images. We imagine a possible practitioner workflow, where the practitioner derives a set of possible hypotheses for the misclassification from \methodname and tests them on the real data. We show further post-hoc simplified examples and a comparison to saliency-based methods in supplementary \secref{suppsec:images-after-training}.



    


\section{Related Work}

Our approach simplifying individual training images builds on \citet{DBLP:conf/iclr/RaghuRKDH21}, where they learn to inject information into the classifier training. Per-instance simplification during training can be seen as a instance of their framework combined with the idea of input simplification. In difference to their methods, \methodname explicitly aims for interpretability through input simplification.

Other interpretability approaches that synthesize inputs include generating counterfactual inputs \citep{DBLP:journals/corr/abs-2103-13701,dombrowski2021diffeomorphic,DBLP:conf/icml/GoyalWEBPL19} or inputs with exaggerated features \citep{DBLP:conf/iclr/SinglaPCB20}. \methodname differs in explicitly optimizing the inputs to be simpler. 

Generative models have often been used in various ways for interpretability such as generating realistic-looking inputs \citep{MONTAVON20181} and by directly training generative classifiers \citep{DBLP:journals/corr/abs-2103-13701,dombrowski2021diffeomorphic}, but we are not aware of any work except \citep{DBLP:journals/corr/abs-2106-10800} (discussed above) to explicitly generate simpler inputs. 
A different approach to reduce input bits while retaining classification performance is to train a compressor that only keeps information that is invariant to predefined label-preserving augmentations. \citet{DBLP:journals/corr/abs-2106-10800} implement this elegant approach in two ways. In their first variant, by training a VAE to reconstruct an unaugmented input from augmented (e.g. rotated, translated, sheared) versions. In their second variant, building on the CLIP \citep{radford2021learning} model, they view images with the same associated text captions as augmented versions of each other. This allows the use of compressed CLIP encodings for classification and achieves up to 1000x compression on Imagenet without decreasing classification accuracy. Their approach focuses  on achieving maximum compression while our approach is focused on interpretability. Their approach requires access to predefined label-preserving augmentations and has reduced classification performance in input space compared to latent/encoding space.




\section{Conclusion}
We propose \textit{SimpleBits}, an information-based method to synthesize simplified inputs. Crucially, \methodname does not require any domain-specific knowledge to constrain or dictate which input components should be removed; instead \methodname itself learns to remove the components of inputs which are least relevant for a given task. 

As an interpretability tool, we show that \methodname is able to remove injected distractors, suggest plausible reasons for misclassification, and recover known radiologic features from condensed datasets. When combined with data condensation, \methodname retains accuracy while greatly reducing the complexity of condensed images.


Our simplification approach sheds light on the amount of information required for a deep network classifier to learn its task. We find that the tradeoff between task performance and input simplification varies by dataset and setting — it is more pronounced for more complex datasets.

\section*{Acknowledgements}
The authors would like to thank Pieter-Jan Kindermans for helping review and providing valuable feedback on early drafts of the paper, and all ML Collective members for useful discussions, feedback and support throughout the project.
Thanks to Aniruddh Raghu and Yann Dubois for open discussions about their related work.
We thank Google Cloud for the GCP Credits Award in support of research in ML Collective.
This work was supported by the German Federal Ministry of Education and Research (BMBF, grant RenormalizedFlows 01IS19077C).

\section*{Reproducibility Statement}
We provide the following information to ensure reproducibility. Main concepts, algorithms and basic architectures are described in Sections \ref{subsec:instance-simplification}, \ref{subsec:post-hoc} and \ref{subsec:dataset-condensation}. Further network architecture details are in Supplementary Sections \ref{suppsec:classifier-network} and \ref{suppsec:simplifier-network}, further optimization details in Supplementary Sections \ref{suppsec:instance-optimization} and
\ref{suppsec:instance-after-alg}. Finally, the code is available under \url{https://tinyurl.com/simple-bits}.

\newpage
\bibliography{main}
\bibliographystyle{icml2022}


\clearpage

\newcommand{\beginsupplement}{%
	\setcounter{table}{0}
	\renewcommand{\thetable}{S\arabic{table}}%
	\setcounter{figure}{0}
	\renewcommand{\thefigure}{S\arabic{figure}}%
	\setcounter{section}{0}
	\renewcommand{\thepage}{S\arabic{page}} 
	\renewcommand{\thesection}{S\arabic{section}}  
	\setcounter{equation}{0}
	\renewcommand{\theequation}{S\arabic{equation}}
}

\appendix
\beginsupplement

\onecolumn
\noindent\makebox[\linewidth]{\rule{\linewidth}{3.5pt}}

\begin{center}
    {\LARGE \bf Supplementary Information for:\\ \titl }
\end{center}
\noindent\makebox[\linewidth]{\rule{\linewidth}{1pt}}

\section*{Supplementary outline}
This document completes the presentation of the main paper with the following:






\newcommand{\specialcell}[2][c]{%
  \begin{tabular}[#1]{@{}c@{}}#2\end{tabular}}

\begin{table}[h]
\begin{center}
\begin{tabular}{c |c |c | p{6.5cm}}
\toprule
\specialcell{\textbf{Supplementary}\\\textbf{Section}}  & \specialcell{\textbf{Type of}\\\textbf{content}} & \specialcell{\textbf{Relevant section}\\ \textbf{in main text}} & \textbf{TL;DR} \\
\midrule
\midrule
\ref{suppsec:bpds-other-models-network} & \textit{Additional experiments} & \secref{subsec:measure_instance_complexity} & Example images with \bpd values produced by other generative models\\
\midrule
\ref{suppsec:architecture-per-instance}, \ref{suppsec:instance-optimization} & \textit{Implementation details} & \secref{subsec:instance-simplification}    & Architecture and optimization details for \emph{simplification during training} \\
\midrule
\ref{suppsec:instance-after-alg} & \textit{Implementation details}  & \secref{subsec:post-hoc} & Optimization details \emph{post-hoc simplification}\\
\midrule

\ref{suppsec:png-storage-space}, \ref{suppsec:retrain-learning-curves}, \ref{suppsec:simplifier-mse-baseline} & \textit{Additional experiments} & \secref{subsec:trade-off-curves} & Baselines, training curves, file size analysis\\
\midrule
\ref{suppsec:images-during-training} & \textit{More figures} & \figref{fig:results-instance-cifar10} & Uncurated sets of simplified images at end of joint training\\
\midrule
\ref{suppsec:spurious-features} & \textit{Additional findings} & \secref{subsec:trade-off-curves} & Potential spurious features revealed by \methodname \\
\midrule
\ref{suppsec:images-condensed} &  \textit{More figures} & \secref{subsec:dataset-condensation} & More simplified condensed dataset images \\
\midrule

\ref{suppsec:continual-learning} &  \textit{Additional experiments} & \secref{subsec:dataset-condensation} & Simplified condensed datasets evaluated in a continual learning setting \\
\midrule

\ref{suppsec:images-after-training} & \textit{Additional findings} & \secref{subsec:post-hoc} & More use cases of \emph{post-training auditing}, with comparison to saliency-based methods \\
\bottomrule

\end{tabular}
\end{center}
\end{table}


\section{BPDs of other generative models}\label{suppsec:bpds-other-models-network}
\begin{figure*}
    \centering
    \includegraphics[width=\linewidth]{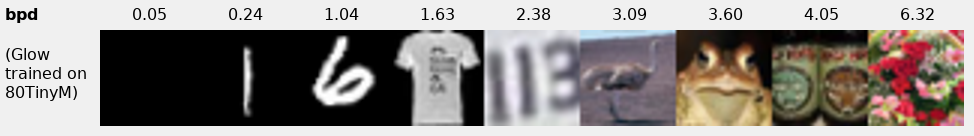}
    \includegraphics[width=\linewidth]{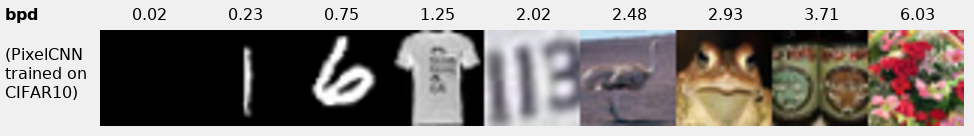}
    \includegraphics[width=\linewidth]{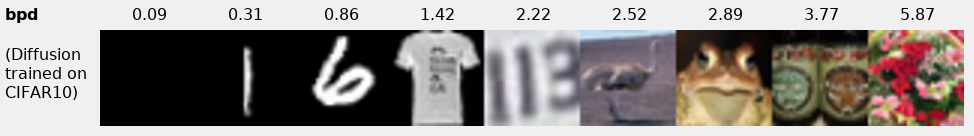}
    \caption{Visualization of the bits-per-dimension ($\mathrm{bpd}$) measure for image complexity, sorted from low to high. Image samples are taken from MNIST, Fashion-MNIST, CIFAR10 and CIFAR100, in addition to a completely black image sample. \bpd is calculated from the density produced by a Glow~\citep{NIPS2018_8224} model pretrained on 80 Million Tiny Images, a PixelCNN model trained on CIFAR10, and a diffusion model trained on CIFAR10.} 
    \label{suppfig:bpds-other-models}
\end{figure*}

\figref{suppfig:bpds-other-models} shows that the bits per dimension produced by other generative models than Glow also correlate well with visual complexity, validating our measure. This is consistent with prior work that found bpds of generative models trained on natural image datasets are strongly influenced by general natural image characteristics independent of any specific dataset \citep{DBLP:conf/nips/KirichenkoIW20,DBLP:conf/nips/SchirrmeisterZB20,DBLP:conf/icml/HavtornFHM21}.

\section{Architecture details for per-instance simplification during training}\label{suppsec:architecture-per-instance}

\subsection{Classifier network}\label{suppsec:classifier-network}
Our classification network is based on the Wide ResNet architecture \citep{DBLP:conf/bmvc/ZagoruykoK16}. We used a version with relatively few parameters with $\textrm{depth}=16$ and $\textrm{widen\_factor}=2$ to allow for fast iteration on experimentation. We used ELU instead of ReLU nonlinearities.

Additionally, we removed batch normalization to avoid interference of normalization layers with the simplification process. We followed the method from \citet{DBLP:conf/icml/BrockDSS21} to create a normalizer-free Wide ResNet. We reparameterize the convolutional layers using Scaled Weight Standardization:
\begin{equation}
    \hat{W_{ij}} = \frac{W_{ij} - \mu_i }{\sqrt{N} \sigma_i },
    \label{eq:scaled_ws}
\end{equation}
where $\mu_i = (1/N) \sum_j W_{ij}$, $\sigma_i^2 = (1/N) \sum_j (W_{ij} - \mu_i)^2$, and $N$ denotes the fan-in.
Further as in \citet{DBLP:conf/icml/BrockDSS21}, "activation functions are also scaled by a non-linearity specific scalar gain $\gamma$, which ensures that the combination of the $\gamma$-scaled activation function and a Scaled Weight Standardized layer is variance preserving." Finally, the output of the residual branch is downscaled by $0.2$, so the function to compute the output becomes $h_{i+1} = h_i + 0.2 \cdot f_i(h_i)$, where $h_i$ denotes the inputs to the $i^{th}$ residual block, and $f_i$ denotes the function computed by the $i^{th}$ residual branch. Unlike \citet{DBLP:conf/icml/BrockDSS21}, we did not multiply scalars $\beta_i$ with the input of the residual branch or learned zero-initialized scalars to multiply with the output of the residual branch, as we did not find these two parts helpful in our setting. We also did not attempt to use Stochastic Depth \citep{DBLP:conf/eccv/HuangSLSW16}, which may further improve upon the accuracies reported here. Due to our small batch sizes (32), we also did not use adaptive gradient clipping.

With this setup, the normalizer-free Wide ResNet with $\textrm{depth}=28$ and $\textrm{widen\_factor}=10$ reached 94.0\% on CIFAR10. To enable faster experiments, we instead use a smaller architecture with $\textrm{depth}=16$ and $\textrm{widen\_factor}=2$ which reached 91.2\% on CIFAR10.

\subsection{Simplifier network}\label{suppsec:simplifier-network}
For the simplifier, we adapted a publicly available implementation of UNet \footnote{\url{https://github.com/junyanz/pytorch-CycleGAN-and-pix2pix}}. We used $\mathrm{num\_down=5}$ downsampling steps, ELU nonlinearities $\mathrm{ngf}=64$ filters in the last conv layer and a simple affine transformation layer instead of a normalization layer. Furthermore, we made the simplifier residual and ensured the output is within $[0,1]$ by adding the output of the UNet to the inverse-sigmoid-transformed input and then reapplying the sigmoid function.

\section{Optimization details for per-instance simplification during training}\label{suppsec:instance-optimization}
First, we note that the single $\mathrm{train\_step}$ helps ensure a correspondence between simplified and original images, and is a technique others have used in meta-learning settings~\citep{DBLP:conf/cvpr/PhamDXL21}.

For stabilizing the optimization of the per-instance simplification during training, we found two further steps helpful. First, we modify:
\begin{align}
\lefteqbox{A}{L_{\mathrm{cls}}}&=l(f'(\bm{X}_\mathrm{orig}), \bm{y}) + l(f(\bm{X}_\mathrm{sim}),\bm{y}) + l(f'(\bm{X}_\mathrm{sim}), \bm{y})
\end{align}
to 
\begin{align}
\lefteqbox{A}{L_{\mathrm{cls}}}&=10 \cdot l(f'(\bm{X}_\mathrm{orig}), \bm{y}) + l(f(\bm{X}_\mathrm{sim}),\bm{y}) + l(f'(\bm{X}_\mathrm{sim}), \bm{y})
\end{align}
as  (a) the gradient magnitudes are much smaller from the losses after unrolling and (b) we want to prioritize the classification loss on the original data.
Additionally, we dynamically turn off $L_{\mathrm{sim}}$ during training, for any example $\bm{x}$ where $l(f'(\bm{x}_\mathrm{orig}), \bm{y})>0.1$, which we also found to further stabilize training.

\section{Optimization details for simplification for auditing post-training}\label{suppsec:instance-after-alg}

\begin{algorithm}[tb!]
\caption{Simplification loss function after training full algorithm}
\footnotesize
\label{alg:post_hoc_full}
\begin{algorithmic}[1]
\STATE{\textbf{given} generative invertible network $G$, input $\bm{x}$ and its latent code $\bm{z}$ (from $G$), simplified input's latent code $\bm{z_{\textrm{sim}}}$, classification network $f$, parameter scaling factors $\bm{s} < 1$}
\STATE{\algcomment{Scale classifier parameters down by $\bm{s}$ to simulate "forgetting"}}
\STATE{$f_{\textrm{scaled}} \leftarrow  \text{ScaleParameters}(f,\bm{s})$} 
\STATE{\algcomment{Sample interpolation factor uniformly between 0 and 1.}}
\STATE{$\alpha \sim U(0,1)$}
\STATE{\algcomment{Interpolate simple and original input in latent space}}
\STATE{$\bm{z}_{\textrm{mixed}}=\alpha\cdot\bm{z}_{sim}+(1-\alpha)\cdot\bm{z}$}
\STATE{\algcomment{Invert  $\bm{z}_{\textrm{mix}}$ using invertible network}}
\STATE{$\bm{x}_{\textrm{mix}}=\textrm{invert}(G,\bm{z}_{\textrm{mix}})$}
\STATE{\algcomment{Predict original and mixed with unscaled classifier}}
\STATE{$\bm{h}=f(\bm{x}),\bm{h}_\textrm{mix}=f(x_\textrm{mix})$}
\STATE{\algcomment{Predict original and mixed with scaled classifier}}
\STATE{$\bm{h}_\textrm{scaled}=f_\textrm{scaled}(\bm{x}),\bm{h}_\textrm{scaled,mix}=f_\textrm{scaled}(\bm{x}_\textrm{mix})$}
\STATE{\algcomment{Compute distance between gradients on scaling factors.}}
\STATE{$L_\textrm{grad}=d\big(\nabla_{\bm{s}} \KL (\bm{h} \Vert \bm{h}_{\mathrm{scaled}}),\nabla_{\bm{s}}\KL (\bm{h} \Vert \bm{h}_{\mathrm{mix,scaled}})\big)$} \label{alg:gradient-distance-full}
\STATE{\algcomment{Compute prediction differences}}
\STATE{$L_\textrm{pred}=\KL (\bm{h} \Vert \bm{h}_{\mathrm{mix}}) + \KL (\bm{h}_{\mathrm{scaled}} \Vert \bm{h}_{\mathrm{mix,scaled}})$} 
\STATE{\algcomment{Invert $\bm{z}_{\textrm{sim}}$ using invertible network}}
\STATE{$\bm{x}_{\textrm{sim}}=\textrm{invert}(G,\bm{z}_{\textrm{sim}})$}
\STATE{\algcomment{Compute needed bits for simplified input}}
\STATE{$L_\textrm{simplification}=-\log p_G(\bm{x}_\mathrm{sim}))$}
\STATE{\algcomment{Return overall loss}}
\STATE{\textbf{return} $L_\textrm{grad} + L_\textrm{pred} + \lambda_\textrm{sim} \cdot L_\textrm{simplification}$ }
\end{algorithmic}
\end{algorithm}

We found it beneficial to optimize the simplified inputs in the latent space of the pre-trained Glow network and to apply our loss functions to all interpolated inputs on the path between original and simple input in latent space instead of only to the simplified input itself. The points on the path include information from the original input and prevent that the optimization is unable to recover some relevant information from the original input. Additionally, we also ensure that the predictions are the same for the original and interpolated inputs for both the original and the scaled model. The complete loss function can be found in Alg. \ref{alg:post_hoc_full}, during training we called it with scaling factors sampled from $s\sim U(0.8,0.95)$ as these values mostly led to similar but more uniformly distributed predictions than the unperturbed network.

\section{PNG-compressed file sizes of simplified images}\label{suppsec:png-storage-space}
\begin{figure*}
    \centering
    \includegraphics[width=0.8\linewidth]{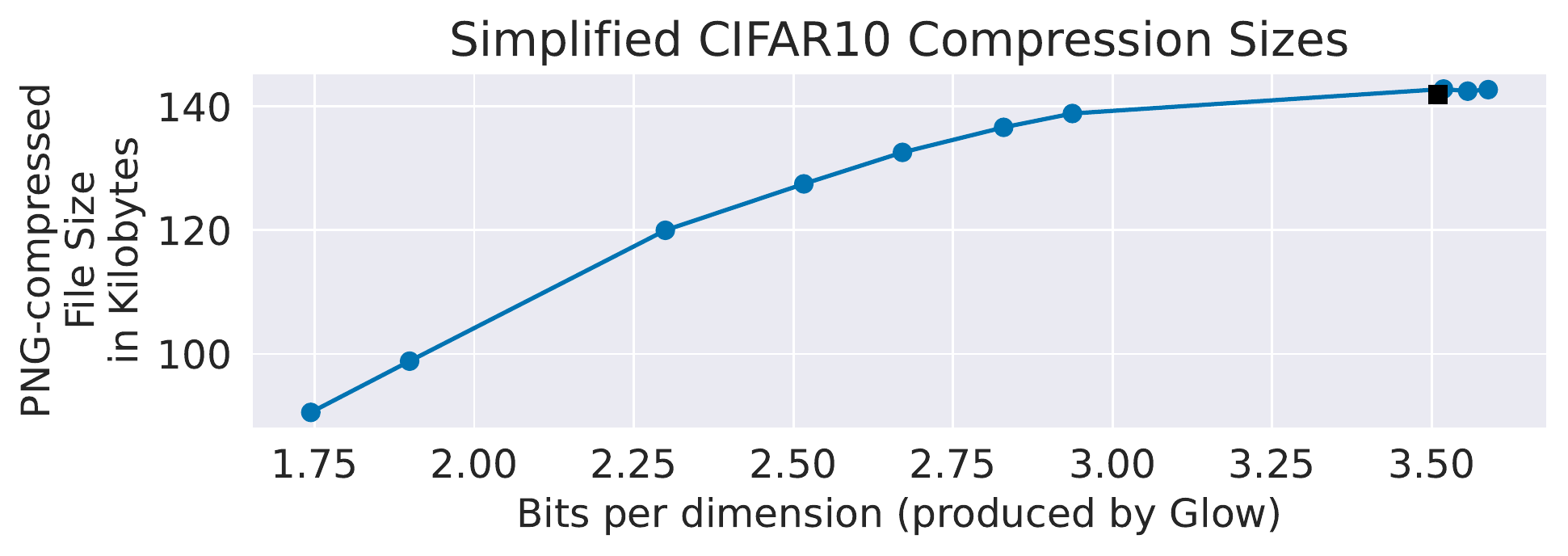}
    \caption{Simplified CIFAR10 images result in smaller storage space when converted into PNG files. Plotted are the average PNG file sizes of the simplified images after joint simplification and classification training (see \secref{subsec:instance-simplification}), against \bpd values produced by Glow, the generative model. Runs with different simplification loss weight $\lambda_\textrm{sim}$ lead to different average file sizes.}
    \label{suppfig:png-bpds-vs-acc}
\end{figure*}

We want to validate that simplified images actually occupy less storage space. We save images in the PNG file format and calculate the file size. 
\figref{suppfig:png-bpds-vs-acc} shows the average PNG file size of the simplified images obtained from our simplification framework.
Concretely, we PNG-compress the images at the end of the joint simplification and classification training and then compute the average file size. Varying the simplification loss weight $\lambda_\textrm{sim}$ leads to different average \bpd values and also different file sizes, with larger $\lambda_\textrm{sim}$ resulting in smaller file sizes. 

\section{Learning Curves for Retraining}\label{suppsec:retrain-learning-curves}
\begin{figure*}
    \centering
    \includegraphics[width=\linewidth]{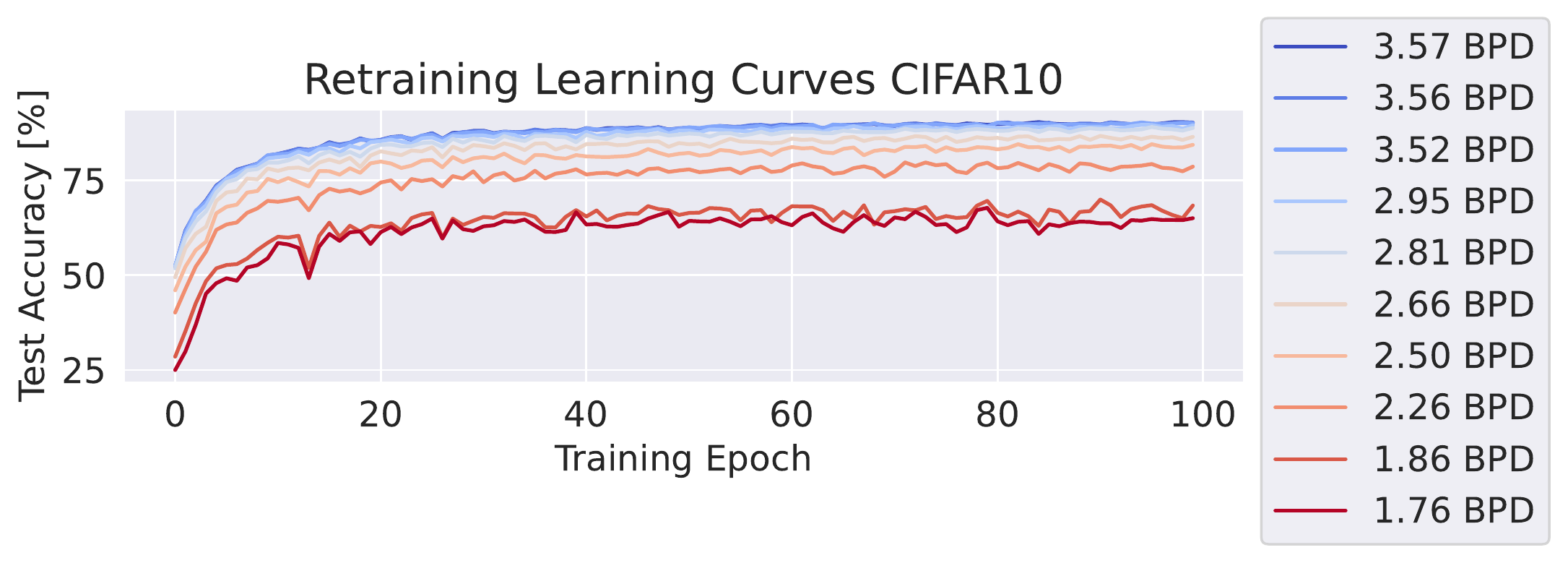}
    \caption{Learning curves for retraining on simplified images on CIFAR10.} 
    \label{suppfig:retraining-simplified-curves}
\end{figure*}

\figref{suppfig:retraining-simplified-curves} shows learning curves during retraining on the simplified images on CIFAR10. There are no noticeable differences in training speed for more or less simplified images.

\section{Simplifier Baselines}\label{suppsec:simplifier-mse-baseline}
\begin{figure*}
    \centering
    \includegraphics[width=\linewidth]{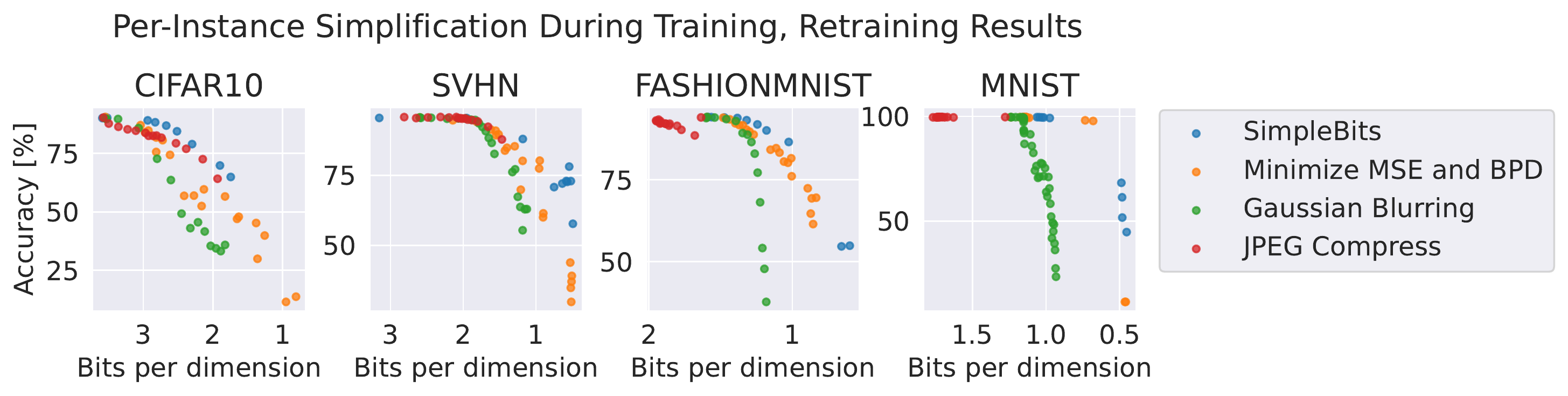}
    \caption{Comparison between  \methodname and two simpler baselines: In the first one, the simplifier network is trained to simultaneously reduce \bpd of the simplified image and the mean squared error between the simplified and the original image. In the second one, gaussian blurring is applied to the input images, different runs vary in the standard deviation used to create the gaussian blurring kernel. In the third one, we use JPEG compression with varying quality levels. Tradeoff curves are worse for the baselines than for \methodname.} 
    \label{suppfig:simplifier-mse-baseline}
\end{figure*}

We implemented three simpler baselines to check whether the losses used in \methodname during training help retain task-relevant information. In the first baseline, we train the simplifier to simultaneously reduce \bpd of the simplified image and the mean squared error between the simplified and the original image. Afterwards we train the classifier on the simplified images and evaluate on the original images the same way as during retraining of \emph{SimpleBits}. In the second baseline, we blur the original images with a gaussian kernel, which also reduces their bpd.  We vary the sigma/standard deviation for the gaussian kernel to trade off smoothness and task-informativeness. In the third baseline, we use lossy JPEG compression with varying quality levels. As in \methodname and the other baselines, we estimate the bits per dimension of the lossy-JPEG-compressed images through our pretrained Glow network for a fair comparison.  The gaussian blurring and JPEG compression each replace the simplifier, so these are  fixed simplifier baselines without training a simplifier. While these three baselines also retain some task-relevant information allowing the classifier to retain above-chance accuracies (see \figref{suppfig:simplifier-mse-baseline}), the tradeoff between \bpd and accuracy is worse than for \emph{SimpleBits}. This shows the losses used in \methodname help retain more task-relevant information compared to these baselines.

\section{More Images Simplified During Training}\label{suppsec:images-during-training}
\begin{figure*}[ht!]
    \centering
    \includegraphics[width=0.7\textwidth]{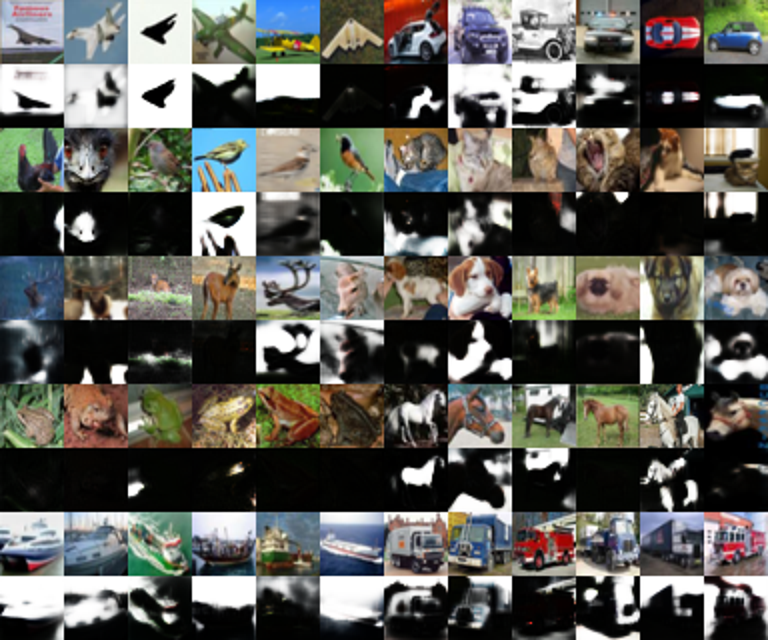}
    \caption{Uncurated set of simplified images with $\lambda_\mathrm{sim}=2.0$, 6 per class.  Rows alternate between original and simplified images.}
    \label{fig:results-instance-uncurated}
\end{figure*}

\begin{figure*}[ht!]
    \centering
    \includegraphics[width=0.7\textwidth]{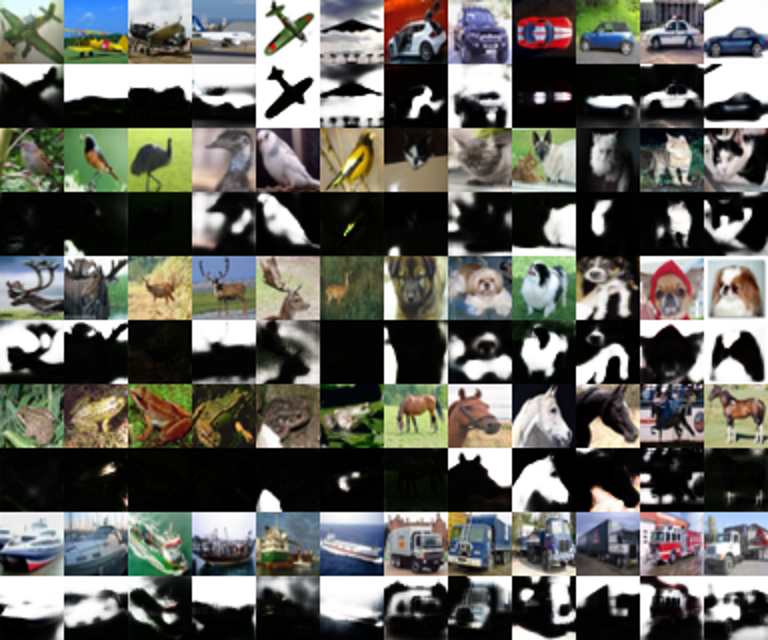}
    \caption{Uncurated set of correctly predicted simplified images with $\lambda_\mathrm{sim}=2.0$, 6 per class. Rows alternate between original and simplified images.}
    \label{fig:results-instance-uncurated-correct}
\end{figure*}

\begin{figure*}[ht!]
    \centering
    \includegraphics[width=0.7\textwidth]{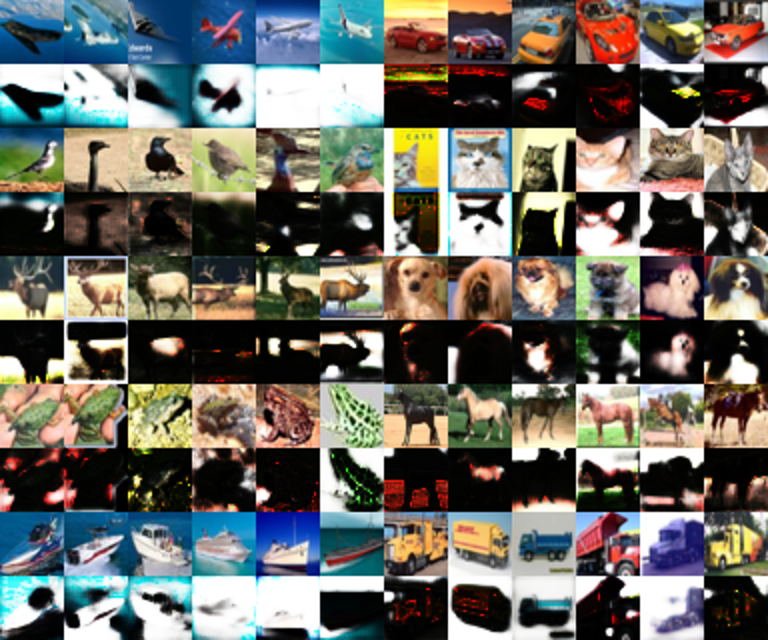}
    \caption{Correctly predicted simplified images with strongest color with $\lambda_\mathrm{sim}=2.0$, 6 per class.}
    \label{fig:results-instance-uncurated-strong-color}
\end{figure*}

We show a larger number of images that were simplified on CIFAR10 during training with the largest simplification loss weight $\lambda_\textrm{sim}=2.0$ in Figures \ref{fig:results-instance-uncurated}, \ref{fig:results-instance-uncurated-correct} and \ref{fig:results-instance-uncurated-strong-color}.


\begin{figure*}[h!]
    \centering
    \includegraphics[width=\linewidth]{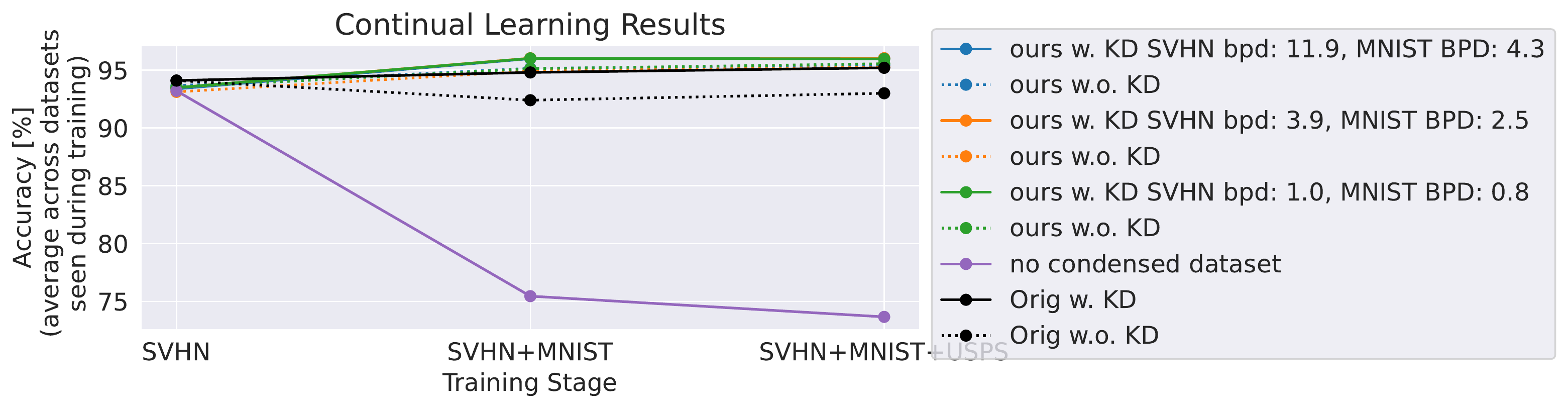}
    \caption{Continual Learning Results without Condensed Dataset (regular sequential training). Conventions as in \figref{fig:results-continual}. Accuracies substantially worse without any condensed dataset.}
    \label{fig:results-continual-no-memory}
\end{figure*}

\begin{figure*}[ht!]
    \centering
    \includegraphics[width=\linewidth]{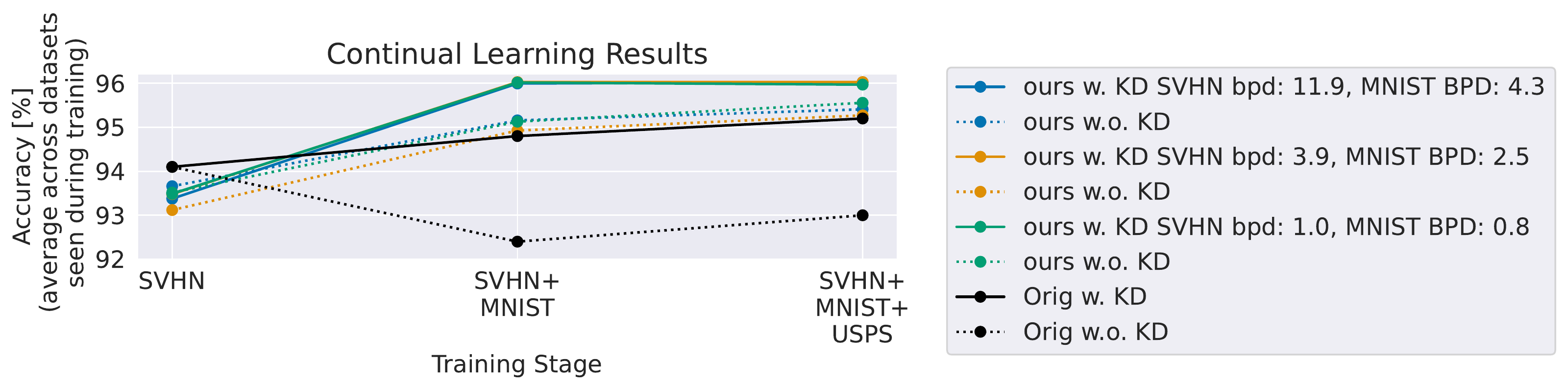}
    \caption{Continual Learning Results. Results for first training on SVHN, then MNIST and then USPS for condensed datasets with varying bits per dimension. Solid lines are with and dashed lines without knowledge distillation. Note that continual learning accuracies remain similar also for substantially reduced bits per dimension. Ablations show that accuracies degrade without any condensed dataset, see supplementary. }
    \label{fig:results-continual}
\end{figure*}

\section{Potentially spurious features uncovered by SimpleBits}\label{suppsec:spurious-features}
\begin{figure*}
    \centering
    \includegraphics[width=0.8\linewidth]{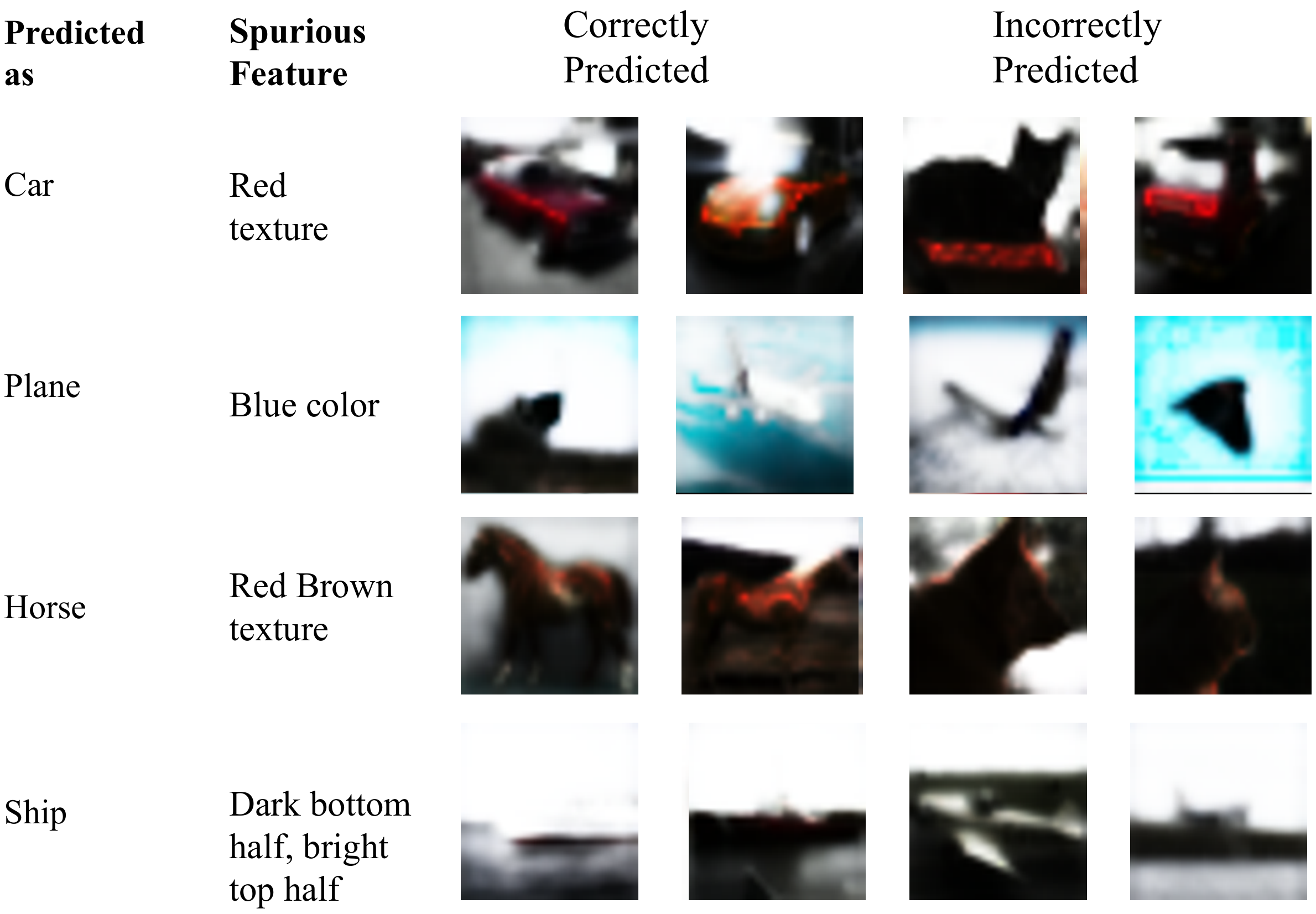}
    \caption{Selected simplified images that highlight potentially spurious features. Two leftmost images are correctly predicted, two rightmost images are incorrectly predicted.} 
    \label{suppfig:spurious-features}
\end{figure*}

\methodname may have the potential to reveal spurious correlations present in the dataset.  We show some simplified images that reveal potentially spurious features in \figref{suppfig:spurious-features}. These observations can be used as a starting point to further investigate whether these features also affect regularly trained classifiers.

\section{More condensed datasets}\label{suppsec:images-condensed}

\begin{figure*}[ht!]
    \centering
    \includegraphics[width=\linewidth]{figures/cifar10condensed.png}
    \includegraphics[width=\linewidth]{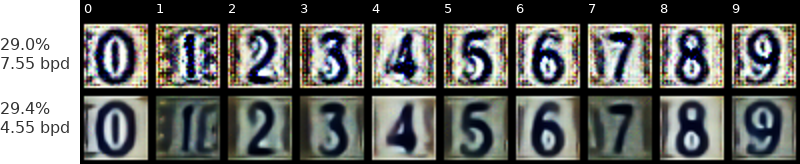}
    \includegraphics[width=\linewidth]{figures/fashion_condensed.png}
    \includegraphics[width=\linewidth]{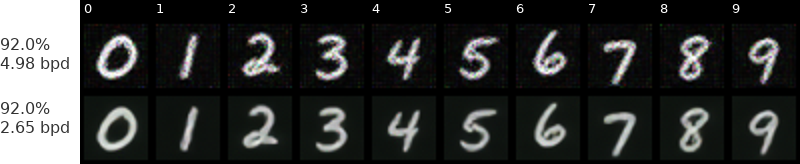}
    \caption{Dataset condensation results with varying simplification loss weight. \textbf{Top:} Individual dots represent accuracies for setting with different simplification loss weights. Accuracies can be retained even with substantially reduced bits per dimension. For 1 image per class, arrows highlight the settings that are visualized below. 
    \textbf{Below:} Condensed datasets with varying simplification loss weight. Per dataset, showing condensed datasets with high (top row) and low (bottom row) bits per dimension. Lower bits per dimension datasets are visually simpler and smoother while mostly retaining accuracies.}
    \label{fig:results-condensation-supp}
\end{figure*}

\begin{figure*}
    \centering
    \includegraphics[width=\linewidth]{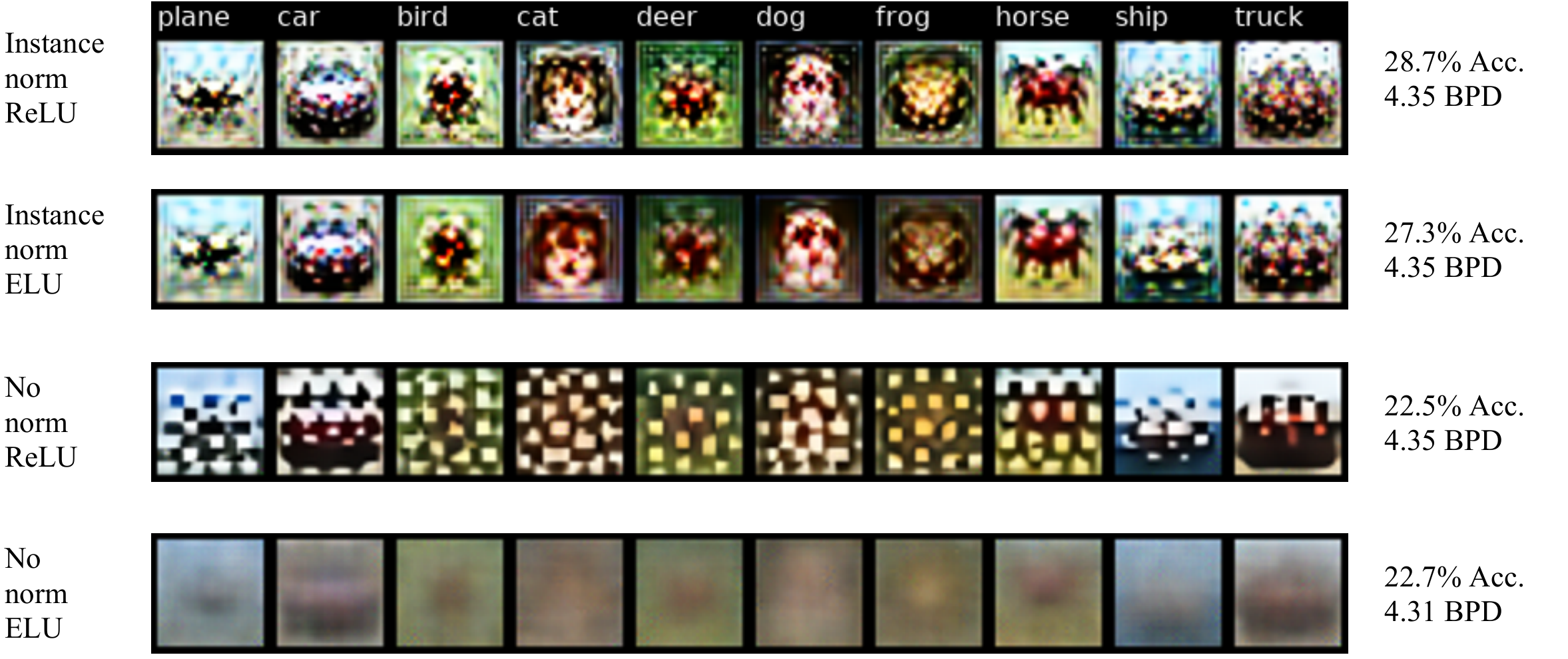}
    \caption{Dataset condensation on CIFAR10 with varying architecture.}
    \label{fig:results-condensation-supp-arch}
\end{figure*}

\begin{figure*}
    \centering
    \includegraphics[width=\linewidth]{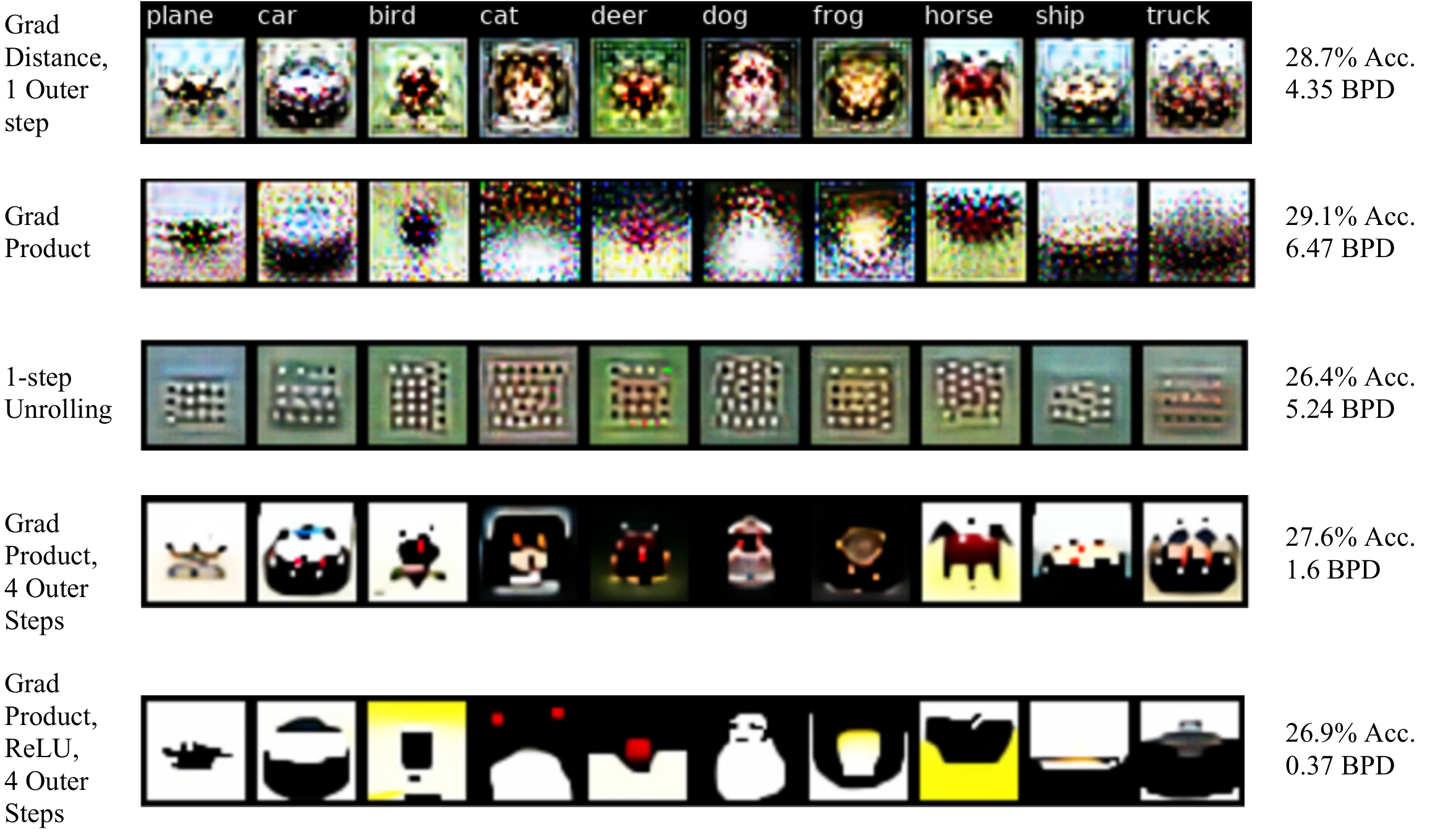}
    \caption{Dataset condensation on CIFAR10 with varying condensation loss and varying outer loop steps, i.e. how many steps the classifier is trained at each training epoch (default 1 in the 1 image per class setting), after each step the condensation loss is again optimized.}
    \label{fig:results-condensation-supp-loss}
\end{figure*}

We also show condensed datasets for MNIST and SVHN (\figref{fig:results-condensation-supp}), some interesting condensed datasets that resulted when we varied the architecture (\figref{fig:results-condensation-supp-arch}) or the condensation loss (from gradient matching to either negative gradient product or single-training-step unrolling, \figref{fig:results-condensation-supp-loss}).

\section{Evaluation of Condensed Datasets for Continual Learning}\label{suppsec:continual-learning}

We also evaluate the simplified condensed datasets in a continual learning setting, following \citep{DBLP:conf/icml/ZhaoB21}. In this task-incremental continual learning setting, the model is trained on different classification datasets sequentially. When training on a new dataset, the model is additionally trained on the condensed versions of the previous datasets.

The continual learning experiment reproduces the setting from \citep{DBLP:conf/icml/ZhaoB21} to first train on SVHN, then on MNIST and finally on USPS \citep{DBLP:journals/pami/Hull94}, using the average accuracy across all three datasets of the classifier at the end of training as the final accuracy (see \citep{DBLP:conf/icml/ZhaoB21} for details).

We created a simpler and faster continual training pipeline that achieves comparable results to \citet{DBLP:conf/icml/ZhaoB21}. First, we train 3 times for 50 epochs on SVHN, with a cosine annealing  learning rate schedule \citep{DBLP:conf/iclr/LoshchilovH17} that is restarted at each time with $lr=0.1$. Then for each MNIST and USPS, we train one cosine annealing cycle of 50 epochs for $lr=0.1$.

We first verified that we can reproduce the prior continual learning results with our simpler training pipeline and find that our training pipeline indeed even slightly outperforms the reported final results (96.0\% vs. 95.2\% with, and 95.4\% vs 93.0\% without knowledge distillation) despite slightly inferior performance in the first training stage (before any continual learning, 93.6\% vs 94.1\%), see following subsection. When using different SVHN and MNIST condensed datasets, we find that we can retain the original continual learning accuracies even with condensed datasets with substantially less ($\sim$9x less) bits per dimension (see Fig.\ref{fig:results-continual}).

\subsection{Without Condensed Dataset}

Our training pipeline still exhibits forgetting when not using any condensed datasets of previously trained-on datasets. As \figref{fig:results-continual-no-memory} shows, the accuracies are far lower than with just regular sequential training. We performed this ablation to ensure forgetting still occurs in our training pipeline.

\section{More images Simplified After Training}\label{suppsec:images-after-training}

\begin{figure*}[ht!]
    \centering
    \includegraphics[width=\linewidth]{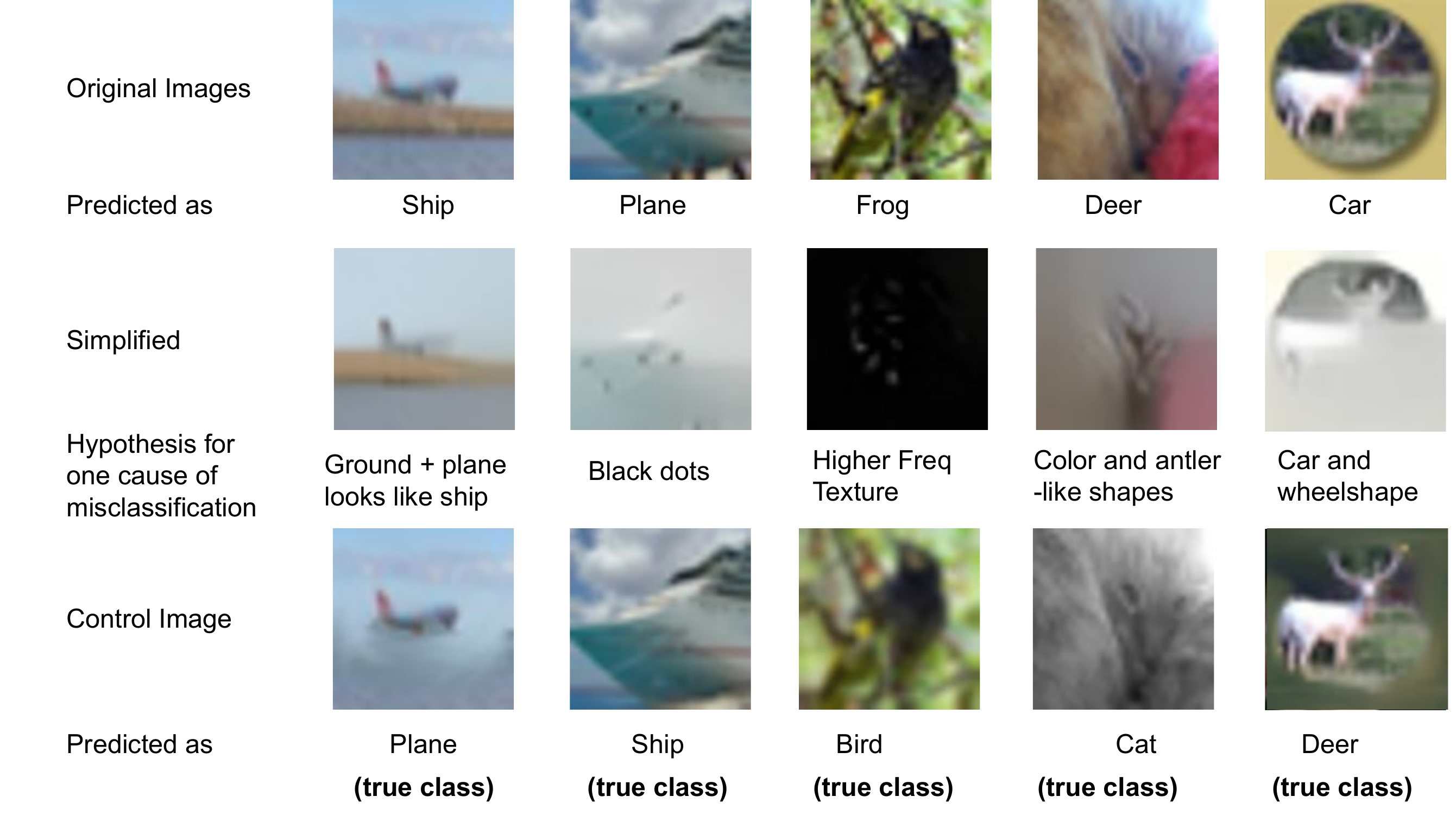}
    \includegraphics[width=\linewidth]{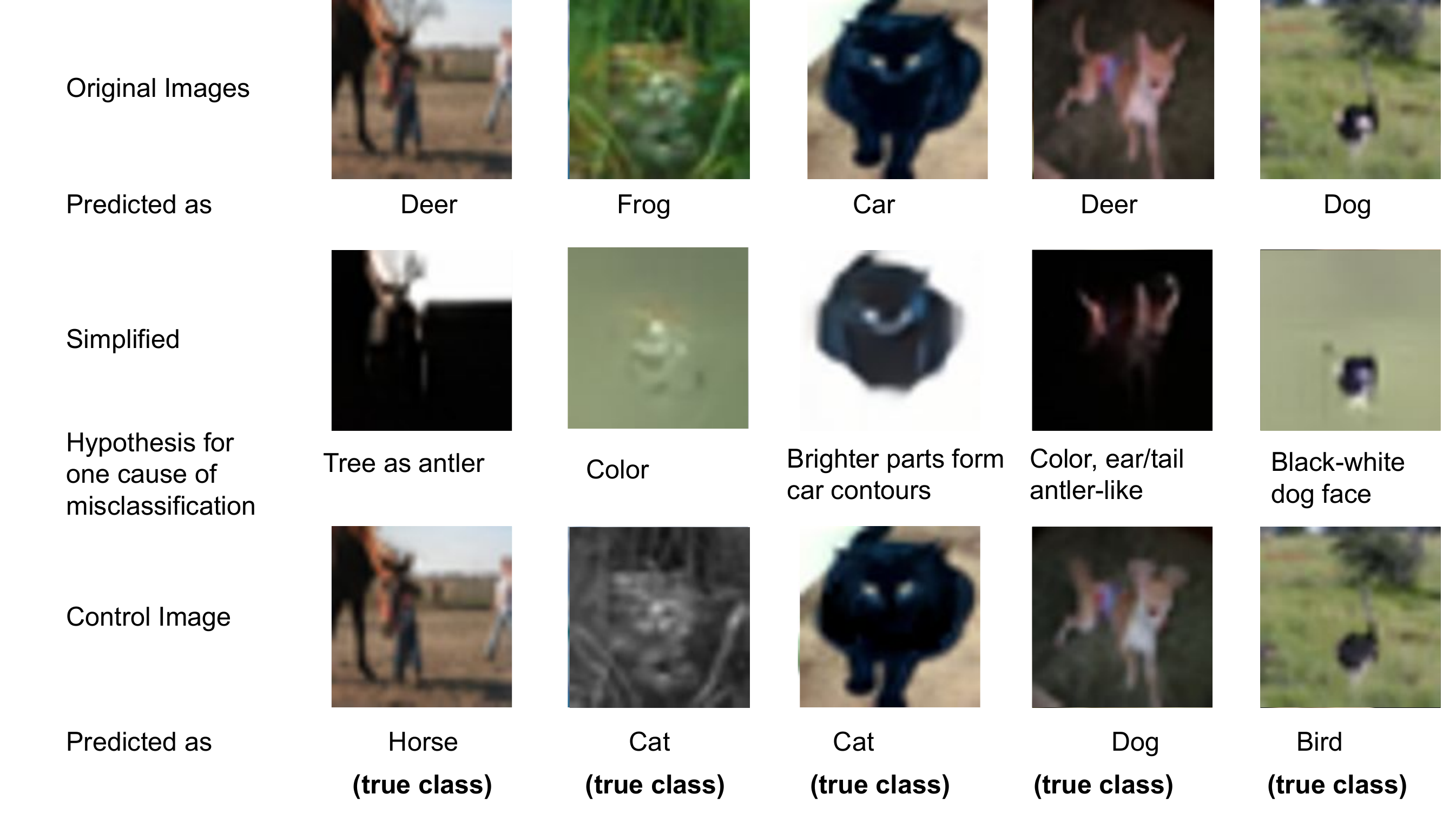}
    \caption{Further examples of post-hoc simplifications of originally misclassified images.}
    \label{fig:results-post-hoc-supp}
\end{figure*}

\begin{figure*}[ht!]
    \centering
    \includegraphics[width=0.8\textwidth]{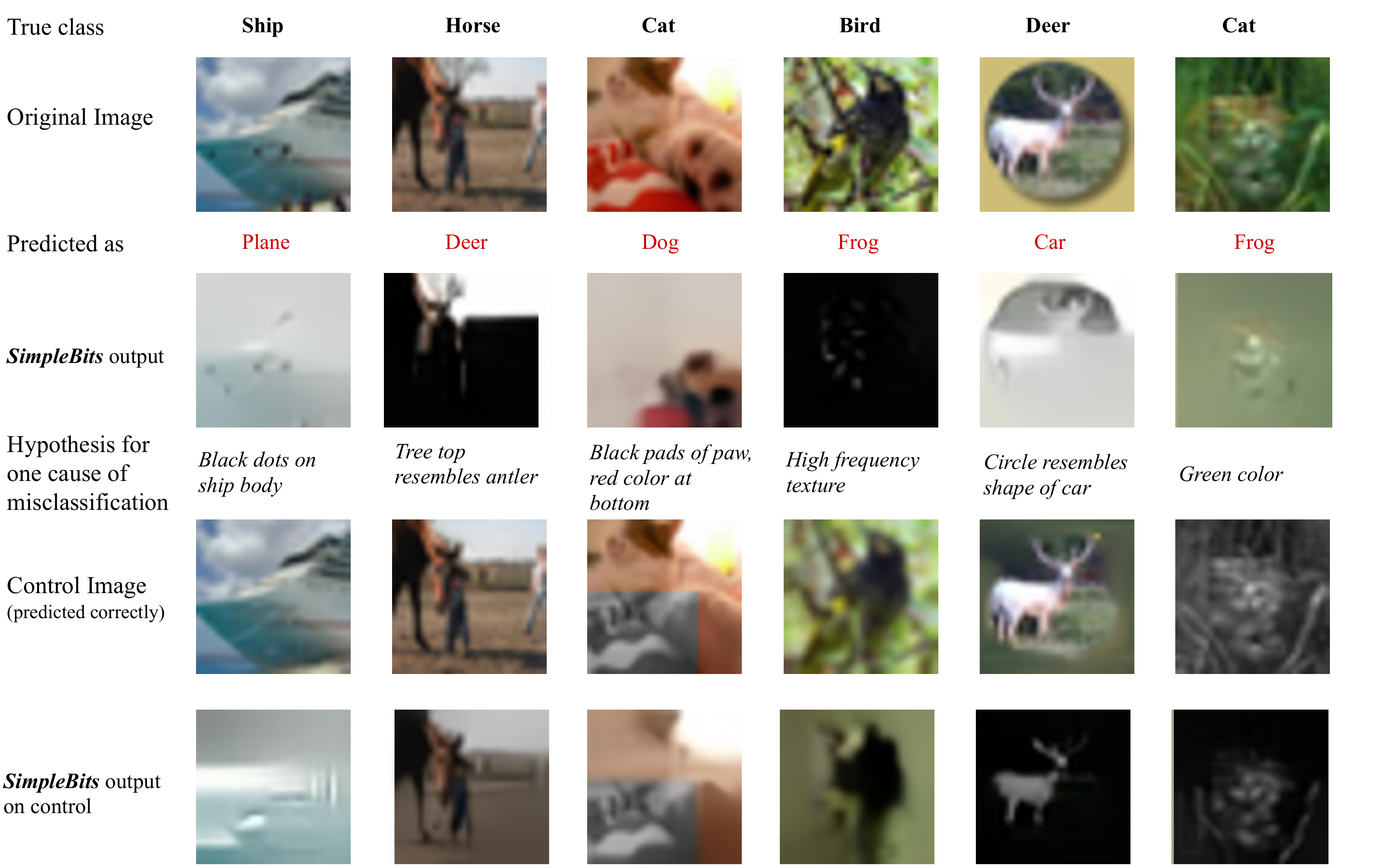}
    \caption{Post-hoc simplifications of control images.}
    \label{suppfig:results-post-hoc-control}
\end{figure*}

\begin{figure*}[ht!]
    \centering
    \includegraphics[width=0.8\textwidth]{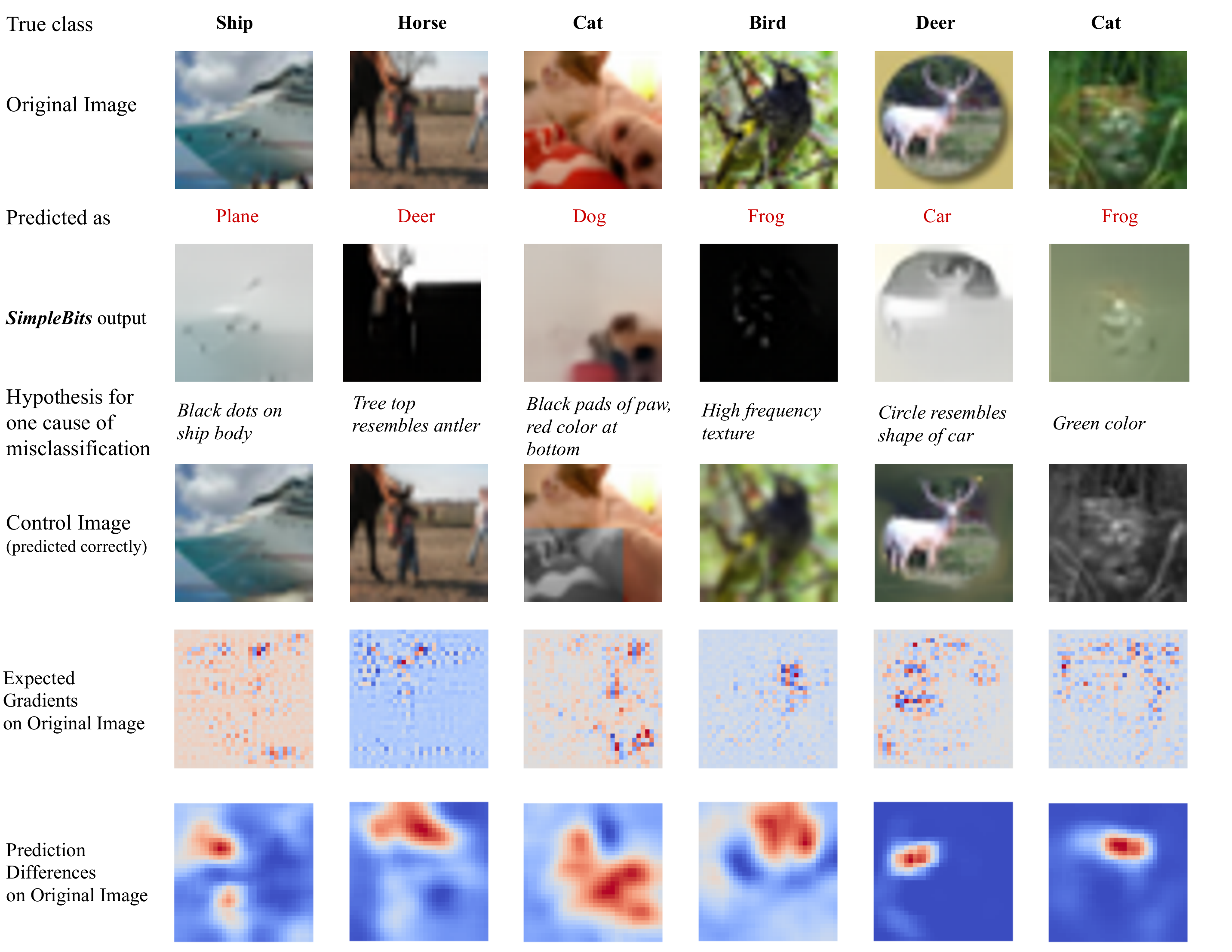}
    \caption{Post-hoc simplifications of misclassified CIFAR-10 examples and saliency maps. In this figure, we also show color-coded saliency maps for expected gradients \citep{erion2021improving} and prediction difference \citep{DBLP:conf/iclr/ZintgrafCAW17} for comparison (red: evidence for and blue: evidence against the predicted class). \methodname reveals more information than saliency methods.
    }
    \label{suppfig:results-post-hoc-saliency}
\end{figure*}

\begin{figure*}[ht!]
    \centering
    \includegraphics[width=0.5\textwidth]{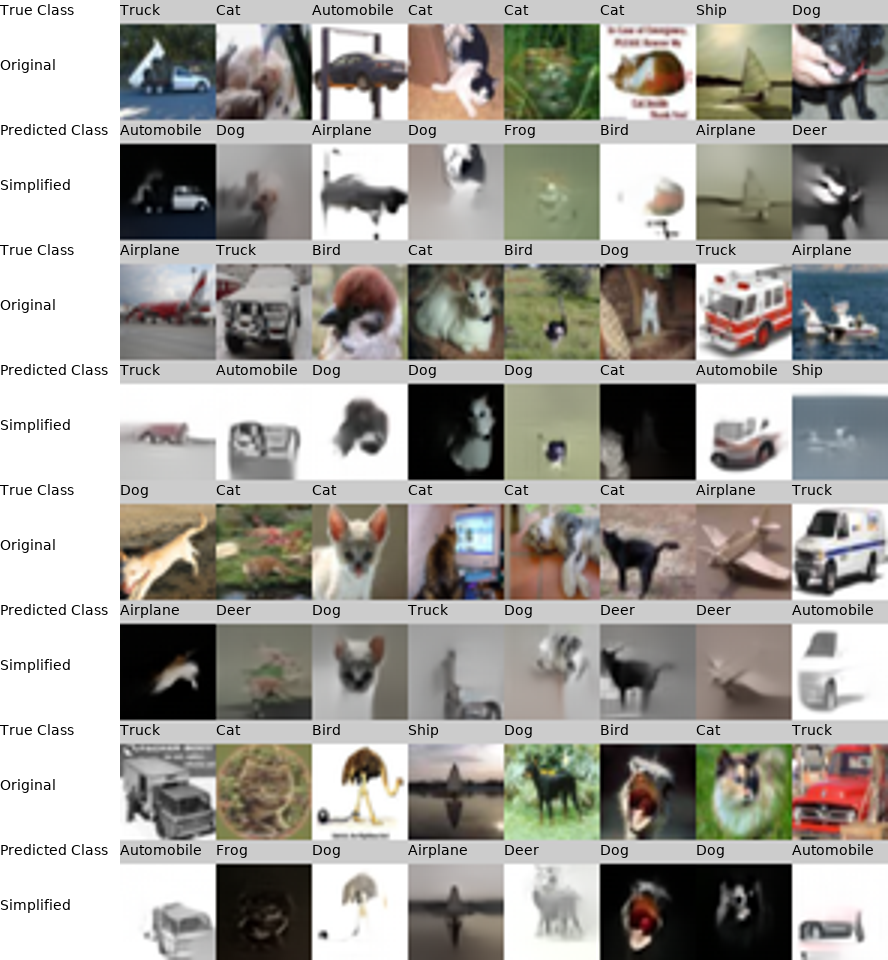}
    \caption{Uncurated post-hoc simplifications of incorrectly predicted images.}
    \label{suppfig:results-post-hoc-uncurated-false-pred}
\end{figure*}

\begin{figure*}[ht!]
    \centering
    \includegraphics[width=0.5\textwidth]{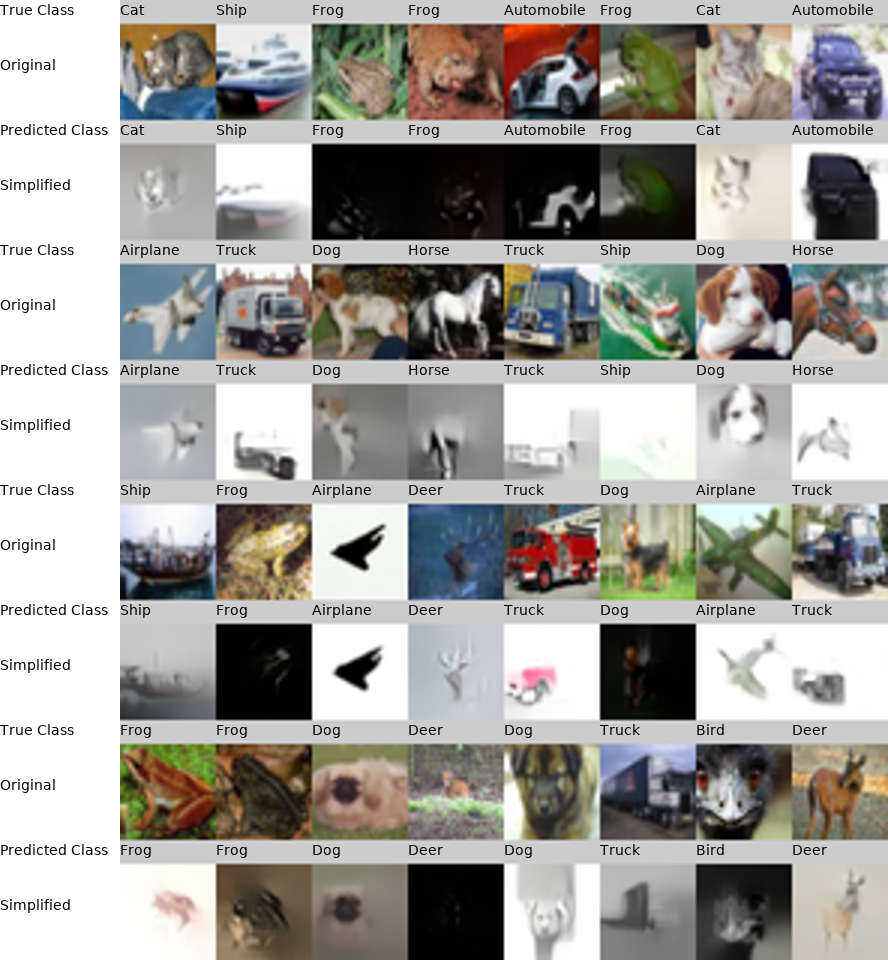}
    \caption{Uncurated post-hoc simplifications of correctly predicted images.}
    \label{suppfig:results-post-hoc-uncurated-true-pred}
\end{figure*}

We show further examples of post-hoc-simplified images for misclassified original images in \figref{fig:results-post-hoc-supp}. We also show the output of \methodname when applied to the control images of \figref{fig:results-post-hoc} in Figure \ref{suppfig:results-post-hoc-control} and a comparison to saliency-based methods in Figure \ref{suppfig:results-post-hoc-saliency}. We also show an uncurated set of incorrectly predicted and correctly predicted images in Figures \ref{suppfig:results-post-hoc-uncurated-false-pred} and \ref{suppfig:results-post-hoc-uncurated-true-pred}.


\end{document}

%% file: math_commands.tex
\usepackage{amsmath,amsfonts,bm}

\def\figref#1{Figure~\ref{#1}}

\def\secref#1{Section~\ref{#1}}

\def\eqref#1{Equation~(\ref{#1})}
\def\Eqref#1{Equation~(\ref{#1})}

\def\1{\bm{1}}

\DeclareMathAlphabet{\mathsfit}{\encodingdefault}{\sfdefault}{m}{sl}
\SetMathAlphabet{\mathsfit}{bold}{\encodingdefault}{\sfdefault}{bx}{n}

\newcommand{\KL}{D_{\mathrm{KL}}}



%% file: main.bbl
\begin{thebibliography}{52}
\providecommand{\natexlab}[1]{#1}
\providecommand{\url}[1]{\texttt{#1}}
\expandafter\ifx\csname urlstyle\endcsname\relax
  \providecommand{\doi}[1]{doi: #1}\else
  \providecommand{\doi}{doi: \begingroup \urlstyle{rm}\Url}\fi

\bibitem[Agarwal et~al.(2021)Agarwal, D'souza, and
  Hooker]{agarwal2021estimating}
Agarwal, C., D'souza, D., and Hooker, S.
\newblock Estimating example difficulty using variance of gradients, 2021.

\bibitem[Banerjee et~al.(2021{\natexlab{a}})Banerjee, Bhimireddy, Burns, Celi,
  Chen, Correa, Dullerud, Ghassemi, Huang, Kuo, Lungren, Palmer, Price,
  Purkayastha, Pyrros, Oakden{-}Rayner, Okechukwu, Seyyed{-}Kalantari, Trivedi,
  Wang, Zaiman, Zhang, and Gichoya]{DBLP:journals/corr/abs-2107-10356}
Banerjee, I., Bhimireddy, A.~R., Burns, J.~L., Celi, L.~A., Chen, L., Correa,
  R., Dullerud, N., Ghassemi, M., Huang, S., Kuo, P., Lungren, M.~P., Palmer,
  L.~J., Price, B.~J., Purkayastha, S., Pyrros, A., Oakden{-}Rayner, L.,
  Okechukwu, C., Seyyed{-}Kalantari, L., Trivedi, H., Wang, R., Zaiman, Z.,
  Zhang, H., and Gichoya, J.~W.
\newblock Reading race: {AI} recognises patient's racial identity in medical
  images.
\newblock \emph{CoRR}, abs/2107.10356, 2021{\natexlab{a}}.
\newblock URL \url{https://arxiv.org/abs/2107.10356}.

\bibitem[Banerjee et~al.(2021{\natexlab{b}})Banerjee, Bhimireddy, Burns, Celi,
  Chen, Correa, Dullerud, Ghassemi, Huang, Kuo, Lungren, Palmer, Price,
  Purkayastha, Pyrros, Oakden-Rayner, Okechukwu, Seyyed-Kalantari, Trivedi,
  Wang, Zaiman, Zhang, and Gichoya]{banerjee2021reading}
Banerjee, I., Bhimireddy, A.~R., Burns, J.~L., Celi, L.~A., Chen, L.-C.,
  Correa, R., Dullerud, N., Ghassemi, M., Huang, S.-C., Kuo, P.-C., Lungren,
  M.~P., Palmer, L., Price, B.~J., Purkayastha, S., Pyrros, A., Oakden-Rayner,
  L., Okechukwu, C., Seyyed-Kalantari, L., Trivedi, H., Wang, R., Zaiman, Z.,
  Zhang, H., and Gichoya, J.~W.
\newblock Reading race: Ai recognises patient's racial identity in medical
  images, 2021{\natexlab{b}}.

\bibitem[Bastings et~al.(2021)Bastings, Ebert, Zablotskaia, Sandholm, and
  Filippova]{bastings2021will}
Bastings, J., Ebert, S., Zablotskaia, P., Sandholm, A., and Filippova, K.
\newblock "will you find these shortcuts?" a protocol for evaluating the
  faithfulness of input salience methods for text classification, 2021.

\bibitem[Bellemare et~al.(2003)Bellemare, Jeanneret, and
  Couture]{bellemare2003sex}
Bellemare, F., Jeanneret, A., and Couture, J.
\newblock Sex differences in thoracic dimensions and configuration.
\newblock \emph{American journal of respiratory and critical care medicine},
  168\penalty0 (3):\penalty0 305--312, 2003.

\bibitem[Brock et~al.(2021)Brock, De, Smith, and
  Simonyan]{DBLP:conf/icml/BrockDSS21}
Brock, A., De, S., Smith, S.~L., and Simonyan, K.
\newblock High-performance large-scale image recognition without normalization.
\newblock In Meila, M. and Zhang, T. (eds.), \emph{Proceedings of the 38th
  International Conference on Machine Learning, {ICML} 2021, 18-24 July 2021,
  Virtual Event}, volume 139 of \emph{Proceedings of Machine Learning
  Research}, pp.\  1059--1071. {PMLR}, 2021.
\newblock URL \url{http://proceedings.mlr.press/v139/brock21a.html}.

\bibitem[Dombrowski et~al.(2021)Dombrowski, Gerken, and
  Kessel]{dombrowski2021diffeomorphic}
Dombrowski, A.-K., Gerken, J.~E., and Kessel, P.
\newblock Diffeomorphic explanations with normalizing flows.
\newblock In \emph{ICML Workshop on Invertible Neural Networks, Normalizing
  Flows, and Explicit Likelihood Models}, 2021.
\newblock URL \url{https://openreview.net/forum?id=ZBR9EpEl6G4}.

\bibitem[D'souza et~al.(2021)D'souza, Nussbaum, Agarwal, and
  Hooker]{dsouza2021tale}
D'souza, D., Nussbaum, Z., Agarwal, C., and Hooker, S.
\newblock A tale of two long tails, 2021.

\bibitem[Dubois et~al.(2021)Dubois, Bloem{-}Reddy, Ullrich, and
  Maddison]{DBLP:journals/corr/abs-2106-10800}
Dubois, Y., Bloem{-}Reddy, B., Ullrich, K., and Maddison, C.~J.
\newblock Lossy compression for lossless prediction.
\newblock \emph{CoRR}, abs/2106.10800, 2021.
\newblock URL \url{https://arxiv.org/abs/2106.10800}.

\bibitem[Erion et~al.(2021)Erion, Janizek, Sturmfels, Lundberg, and
  Lee]{erion2021improving}
Erion, G., Janizek, J.~D., Sturmfels, P., Lundberg, S.~M., and Lee, S.-I.
\newblock Improving performance of deep learning models with axiomatic
  attribution priors and expected gradients.
\newblock \emph{Nature Machine Intelligence}, pp.\  1--12, 2021.

\bibitem[Finn et~al.(2017)Finn, Abbeel, and Levine]{DBLP:conf/icml/FinnAL17}
Finn, C., Abbeel, P., and Levine, S.
\newblock Model-agnostic meta-learning for fast adaptation of deep networks.
\newblock In Precup, D. and Teh, Y.~W. (eds.), \emph{Proceedings of the 34th
  International Conference on Machine Learning, {ICML} 2017, Sydney, NSW,
  Australia, 6-11 August 2017}, volume~70 of \emph{Proceedings of Machine
  Learning Research}, pp.\  1126--1135. {PMLR}, 2017.
\newblock URL \url{http://proceedings.mlr.press/v70/finn17a.html}.

\bibitem[Geirhos et~al.(2020)Geirhos, Jacobsen, Michaelis, Zemel, Brendel,
  Bethge, and Wichmann]{geirhos2020shortcut}
Geirhos, R., Jacobsen, J.-H., Michaelis, C., Zemel, R., Brendel, W., Bethge,
  M., and Wichmann, F.~A.
\newblock Shortcut learning in deep neural networks.
\newblock \emph{Nature Machine Intelligence}, 2\penalty0 (11):\penalty0
  665--673, 2020.

\bibitem[Goyal et~al.(2019)Goyal, Wu, Ernst, Batra, Parikh, and
  Lee]{DBLP:conf/icml/GoyalWEBPL19}
Goyal, Y., Wu, Z., Ernst, J., Batra, D., Parikh, D., and Lee, S.
\newblock Counterfactual visual explanations.
\newblock In Chaudhuri, K. and Salakhutdinov, R. (eds.), \emph{Proceedings of
  the 36th International Conference on Machine Learning, {ICML} 2019, 9-15 June
  2019, Long Beach, California, {USA}}, volume~97 of \emph{Proceedings of
  Machine Learning Research}, pp.\  2376--2384. {PMLR}, 2019.
\newblock URL \url{http://proceedings.mlr.press/v97/goyal19a.html}.

\bibitem[Havtorn et~al.(2021)Havtorn, Frellsen, Hauberg, and
  Maal{\o}e]{DBLP:conf/icml/HavtornFHM21}
Havtorn, J.~D., Frellsen, J., Hauberg, S., and Maal{\o}e, L.
\newblock Hierarchical vaes know what they don't know.
\newblock In Meila, M. and Zhang, T. (eds.), \emph{Proceedings of the 38th
  International Conference on Machine Learning, {ICML} 2021, 18-24 July 2021,
  Virtual Event}, volume 139 of \emph{Proceedings of Machine Learning
  Research}, pp.\  4117--4128. {PMLR}, 2021.
\newblock URL \url{http://proceedings.mlr.press/v139/havtorn21a.html}.

\bibitem[Hinton \& van Camp(1993)Hinton and van Camp]{DBLP:conf/colt/HintonC93}
Hinton, G.~E. and van Camp, D.
\newblock Keeping the neural networks simple by minimizing the description
  length of the weights.
\newblock In Pitt, L. (ed.), \emph{Proceedings of the Sixth Annual {ACM}
  Conference on Computational Learning Theory, {COLT} 1993, Santa Cruz, CA,
  USA, July 26-28, 1993}, pp.\  5--13. {ACM}, 1993.
\newblock \doi{10.1145/168304.168306}.
\newblock URL \url{https://doi.org/10.1145/168304.168306}.

\bibitem[Hooker et~al.(2019)Hooker, Erhan, Kindermans, and
  Kim]{NEURIPS2019_fe4b8556}
Hooker, S., Erhan, D., Kindermans, P.-J., and Kim, B.
\newblock A benchmark for interpretability methods in deep neural networks.
\newblock In Wallach, H., Larochelle, H., Beygelzimer, A., d\' Alch\'{e}-Buc,
  F., Fox, E., and Garnett, R. (eds.), \emph{Advances in Neural Information
  Processing Systems}, volume~32. Curran Associates, Inc., 2019.
\newblock URL
  \url{https://proceedings.neurips.cc/paper/2019/file/fe4b8556000d0f0cae99daa5c5c5a410-Paper.pdf}.

\bibitem[Huang et~al.(2016)Huang, Sun, Liu, Sedra, and
  Weinberger]{DBLP:conf/eccv/HuangSLSW16}
Huang, G., Sun, Y., Liu, Z., Sedra, D., and Weinberger, K.~Q.
\newblock Deep networks with stochastic depth.
\newblock In Leibe, B., Matas, J., Sebe, N., and Welling, M. (eds.),
  \emph{Computer Vision - {ECCV} 2016 - 14th European Conference, Amsterdam,
  The Netherlands, October 11-14, 2016, Proceedings, Part {IV}}, volume 9908 of
  \emph{Lecture Notes in Computer Science}, pp.\  646--661. Springer, 2016.
\newblock \doi{10.1007/978-3-319-46493-0\_39}.
\newblock URL \url{https://doi.org/10.1007/978-3-319-46493-0\_39}.

\bibitem[Hull(1994)]{DBLP:journals/pami/Hull94}
Hull, J.~J.
\newblock A database for handwritten text recognition research.
\newblock \emph{{IEEE} Trans. Pattern Anal. Mach. Intell.}, 16\penalty0
  (5):\penalty0 550--554, 1994.
\newblock \doi{10.1109/34.291440}.
\newblock URL \url{https://doi.org/10.1109/34.291440}.

\bibitem[Hvilsh{\o}j et~al.(2021)Hvilsh{\o}j, Iosifidis, and
  Assent]{DBLP:journals/corr/abs-2103-13701}
Hvilsh{\o}j, F., Iosifidis, A., and Assent, I.
\newblock {ECINN:} efficient counterfactuals from invertible neural networks.
\newblock \emph{CoRR}, abs/2103.13701, 2021.
\newblock URL \url{https://arxiv.org/abs/2103.13701}.

\bibitem[Jany \& Welte(2019)Jany and Welte]{jany2019pleural}
Jany, B. and Welte, T.
\newblock Pleural effusion in adults—etiology, diagnosis, and treatment.
\newblock \emph{Deutsches {\"A}rzteblatt International}, 116\penalty0
  (21):\penalty0 377, 2019.

\bibitem[Johnson et~al.(2019{\natexlab{a}})Johnson, Pollard, Berkowitz,
  Greenbaum, Lungren, Deng, Mark, and Horng]{johnson2019mimic}
Johnson, A.~E., Pollard, T.~J., Berkowitz, S.~J., Greenbaum, N.~R., Lungren,
  M.~P., Deng, C.-y., Mark, R.~G., and Horng, S.
\newblock Mimic-cxr, a de-identified publicly available database of chest
  radiographs with free-text reports.
\newblock \emph{Scientific data}, 6\penalty0 (1):\penalty0 1--8,
  2019{\natexlab{a}}.

\bibitem[Johnson et~al.(2019{\natexlab{b}})Johnson, Pollard, Greenbaum,
  Lungren, Deng, Peng, Lu, Mark, Berkowitz, and Horng]{johnson2019mimicjpg}
Johnson, A.~E., Pollard, T.~J., Greenbaum, N.~R., Lungren, M.~P., Deng, C.-y.,
  Peng, Y., Lu, Z., Mark, R.~G., Berkowitz, S.~J., and Horng, S.
\newblock Mimic-cxr-jpg, a large publicly available database of labeled chest
  radiographs.
\newblock \emph{arXiv preprint arXiv:1901.07042}, 2019{\natexlab{b}}.

\bibitem[Kingma \& Ba(2015)Kingma and Ba]{DBLP:journals/corr/KingmaB14}
Kingma, D.~P. and Ba, J.
\newblock Adam: {A} method for stochastic optimization.
\newblock In Bengio, Y. and LeCun, Y. (eds.), \emph{3rd International
  Conference on Learning Representations, {ICLR} 2015, San Diego, CA, USA, May
  7-9, 2015, Conference Track Proceedings}, 2015.
\newblock URL \url{http://arxiv.org/abs/1412.6980}.

\bibitem[Kingma \& Dhariwal(2018)Kingma and Dhariwal]{NIPS2018_8224}
Kingma, D.~P. and Dhariwal, P.
\newblock Glow: Generative flow with invertible 1x1 convolutions.
\newblock In Bengio, S., Wallach, H., Larochelle, H., Grauman, K.,
  Cesa-Bianchi, N., and Garnett, R. (eds.), \emph{Advances in Neural
  Information Processing Systems (NeuRIPs)}, pp.\  10215--10224, 2018.

\bibitem[Kirichenko et~al.(2020)Kirichenko, Izmailov, and
  Wilson]{DBLP:conf/nips/KirichenkoIW20}
Kirichenko, P., Izmailov, P., and Wilson, A.~G.
\newblock Why normalizing flows fail to detect out-of-distribution data.
\newblock In Larochelle, H., Ranzato, M., Hadsell, R., Balcan, M., and Lin, H.
  (eds.), \emph{Advances in Neural Information Processing Systems 33: Annual
  Conference on Neural Information Processing Systems 2020, NeurIPS 2020,
  December 6-12, 2020, virtual}, 2020.
\newblock URL
  \url{https://proceedings.neurips.cc/paper/2020/hash/ecb9fe2fbb99c31f567e9823e884dbec-Abstract.html}.

\bibitem[Krizhevsky(2009)]{Cifar10_Krizhevsky09learningmultiple}
Krizhevsky, A.
\newblock Learning multiple layers of features from tiny images.
\newblock Technical report, 2009.

\bibitem[LeCun \& Cortes(2010)LeCun and
  Cortes]{lecun-mnisthandwrittendigit-2010}
LeCun, Y. and Cortes, C.
\newblock {MNIST} handwritten digit database.
\newblock 2010.
\newblock URL \url{http://yann.lecun.com/exdb/mnist/}.

\bibitem[Loshchilov \& Hutter(2017)Loshchilov and
  Hutter]{DBLP:conf/iclr/LoshchilovH17}
Loshchilov, I. and Hutter, F.
\newblock {SGDR:} stochastic gradient descent with warm restarts.
\newblock In \emph{5th International Conference on Learning Representations,
  {ICLR} 2017, Toulon, France, April 24-26, 2017, Conference Track
  Proceedings}. OpenReview.net, 2017.
\newblock URL \url{https://openreview.net/forum?id=Skq89Scxx}.

\bibitem[Loshchilov \& Hutter(2019)Loshchilov and
  Hutter]{DBLP:conf/iclr/LoshchilovH19}
Loshchilov, I. and Hutter, F.
\newblock Decoupled weight decay regularization.
\newblock In \emph{7th International Conference on Learning Representations,
  {ICLR} 2019, New Orleans, LA, USA, May 6-9, 2019}. OpenReview.net, 2019.
\newblock URL \url{https://openreview.net/forum?id=Bkg6RiCqY7}.

\bibitem[Maclaurin et~al.(2015)Maclaurin, Duvenaud, and
  Adams]{pmlr-v37-maclaurin15}
Maclaurin, D., Duvenaud, D., and Adams, R.
\newblock Gradient-based hyperparameter optimization through reversible
  learning.
\newblock In Bach, F. and Blei, D. (eds.), \emph{Proceedings of the 32nd
  International Conference on Machine Learning}, volume~37 of \emph{Proceedings
  of Machine Learning Research}, pp.\  2113--2122, Lille, France, 07--09 Jul
  2015. PMLR.
\newblock URL \url{https://proceedings.mlr.press/v37/maclaurin15.html}.

\bibitem[Madsen et~al.(2021)Madsen, Meade, Adlakha, and
  Reddy]{madsen2021evaluating}
Madsen, A., Meade, N., Adlakha, V., and Reddy, S.
\newblock Evaluating the faithfulness of importance measures in nlp by
  recursively masking allegedly important tokens and retraining, 2021.

\bibitem[Makino et~al.(2020)Makino, Jastrzebski, Oleszkiewicz, Chacko,
  Ehrenpreis, Samreen, Chhor, Kim, Lee, Pysarenko,
  et~al.]{makino2020differences}
Makino, T., Jastrzebski, S., Oleszkiewicz, W., Chacko, C., Ehrenpreis, R.,
  Samreen, N., Chhor, C., Kim, E., Lee, J., Pysarenko, K., et~al.
\newblock Differences between human and machine perception in medical
  diagnosis.
\newblock \emph{arXiv preprint arXiv:2011.14036}, 2020.

\bibitem[Montavon et~al.(2018)Montavon, Samek, and Müller]{MONTAVON20181}
Montavon, G., Samek, W., and Müller, K.-R.
\newblock Methods for interpreting and understanding deep neural networks.
\newblock \emph{Digital Signal Processing}, 73:\penalty0 1--15, 2018.
\newblock ISSN 1051-2004.
\newblock \doi{https://doi.org/10.1016/j.dsp.2017.10.011}.
\newblock URL
  \url{https://www.sciencedirect.com/science/article/pii/S1051200417302385}.

\bibitem[Netzer et~al.(2011)Netzer, Wang, Coates, Bissacco, Wu, and
  Ng]{Netzer_SVHN}
Netzer, Y., Wang, T., Coates, A., Bissacco, A., Wu, B., and Ng, A.~Y.
\newblock Reading digits in natural images with unsupervised feature learning.
\newblock In \emph{NIPS Workshop on Deep Learning and Unsupervised Feature
  Learning}, 2011.

\bibitem[Nguyen et~al.(2021)Nguyen, Novak, Xiao, and
  Lee]{DBLP:journals/corr/abs-2107-13034}
Nguyen, T., Novak, R., Xiao, L., and Lee, J.
\newblock Dataset distillation with infinitely wide convolutional networks.
\newblock \emph{CoRR}, abs/2107.13034, 2021.
\newblock URL \url{https://arxiv.org/abs/2107.13034}.

\bibitem[Pham et~al.(2021)Pham, Dai, Xie, and Le]{DBLP:conf/cvpr/PhamDXL21}
Pham, H., Dai, Z., Xie, Q., and Le, Q.~V.
\newblock Meta pseudo labels.
\newblock In \emph{{IEEE} Conference on Computer Vision and Pattern
  Recognition, {CVPR} 2021, virtual, June 19-25, 2021}, pp.\  11557--11568.
  Computer Vision Foundation / {IEEE}, 2021.
\newblock URL
  \url{https://openaccess.thecvf.com/content/CVPR2021/html/Pham\_Meta\_Pseudo\_Labels\_CVPR\_2021\_paper.html}.

\bibitem[Raasch et~al.(1982)Raasch, Carsky, Lane, O'Callaghan, and
  Heitzman]{raasch1982pleural}
Raasch, B., Carsky, E., Lane, E., O'Callaghan, J., and Heitzman, E.
\newblock Pleural effusion: explanation of some typical appearances.
\newblock \emph{American Journal of Roentgenology}, 139\penalty0 (5):\penalty0
  899--904, 1982.

\bibitem[Radford et~al.(2021)Radford, Kim, Hallacy, Ramesh, Goh, Agarwal,
  Sastry, Askell, Mishkin, Clark, et~al.]{radford2021learning}
Radford, A., Kim, J.~W., Hallacy, C., Ramesh, A., Goh, G., Agarwal, S., Sastry,
  G., Askell, A., Mishkin, P., Clark, J., et~al.
\newblock Learning transferable visual models from natural language
  supervision.
\newblock \emph{arXiv preprint arXiv:2103.00020}, 2021.

\bibitem[Raghu et~al.(2021)Raghu, Raghu, Kornblith, Duvenaud, and
  Hinton]{DBLP:conf/iclr/RaghuRKDH21}
Raghu, A., Raghu, M., Kornblith, S., Duvenaud, D., and Hinton, G.~E.
\newblock Teaching with commentaries.
\newblock In \emph{9th International Conference on Learning Representations,
  {ICLR} 2021, Virtual Event, Austria, May 3-7, 2021}. OpenReview.net, 2021.
\newblock URL \url{https://openreview.net/forum?id=4RbdgBh9gE}.

\bibitem[Raghu \& Schmidt(2020)Raghu and Schmidt]{raghu2020survey}
Raghu, M. and Schmidt, E.
\newblock A survey of deep learning for scientific discovery.
\newblock \emph{arXiv preprint arXiv:2003.11755}, 2020.

\bibitem[Ronneberger et~al.(2015)Ronneberger, Fischer, and
  Brox]{DBLP:conf/miccai/RonnebergerFB15}
Ronneberger, O., Fischer, P., and Brox, T.
\newblock U-net: Convolutional networks for biomedical image segmentation.
\newblock In Navab, N., Hornegger, J., III, W. M.~W., and Frangi, A.~F. (eds.),
  \emph{Medical Image Computing and Computer-Assisted Intervention - {MICCAI}
  2015 - 18th International Conference Munich, Germany, October 5 - 9, 2015,
  Proceedings, Part {III}}, volume 9351 of \emph{Lecture Notes in Computer
  Science}, pp.\  234--241. Springer, 2015.
\newblock \doi{10.1007/978-3-319-24574-4\_28}.
\newblock URL \url{https://doi.org/10.1007/978-3-319-24574-4\_28}.

\bibitem[Schirrmeister et~al.(2020)Schirrmeister, Zhou, Ball, and
  Zhang]{DBLP:conf/nips/SchirrmeisterZB20}
Schirrmeister, R., Zhou, Y., Ball, T., and Zhang, D.
\newblock Understanding anomaly detection with deep invertible networks through
  hierarchies of distributions and features.
\newblock In Larochelle, H., Ranzato, M., Hadsell, R., Balcan, M., and Lin, H.
  (eds.), \emph{Advances in Neural Information Processing Systems 33: Annual
  Conference on Neural Information Processing Systems 2020, NeurIPS 2020,
  December 6-12, 2020, virtual}, 2020.
\newblock URL
  \url{https://proceedings.neurips.cc/paper/2020/hash/f106b7f99d2cb30c3db1c3cc0fde9ccb-Abstract.html}.

\bibitem[Serr{\`{a}} et~al.(2020)Serr{\`{a}}, {\'{A}}lvarez, G{\'{o}}mez,
  Slizovskaia, N{\'{u}}{\~{n}}ez, and Luque]{DBLP:conf/iclr/SerraAGSNL20}
Serr{\`{a}}, J., {\'{A}}lvarez, D., G{\'{o}}mez, V., Slizovskaia, O.,
  N{\'{u}}{\~{n}}ez, J.~F., and Luque, J.
\newblock Input complexity and out-of-distribution detection with
  likelihood-based generative models.
\newblock In \emph{8th International Conference on Learning Representations,
  {ICLR} 2020, Addis Ababa, Ethiopia, April 26-30, 2020}. OpenReview.net, 2020.
\newblock URL \url{https://openreview.net/forum?id=SyxIWpVYvr}.

\bibitem[Shannon(1948)]{Shannon1948}
Shannon, C.~E.
\newblock A mathematical theory of communication.
\newblock \emph{The Bell System Technical Journal}, 27\penalty0 (3):\penalty0
  379--423, July 1948.

\bibitem[Singla et~al.(2020)Singla, Pollack, Chen, and
  Batmanghelich]{DBLP:conf/iclr/SinglaPCB20}
Singla, S., Pollack, B., Chen, J., and Batmanghelich, K.
\newblock Explanation by progressive exaggeration.
\newblock In \emph{8th International Conference on Learning Representations,
  {ICLR} 2020, Addis Ababa, Ethiopia, April 26-30, 2020}. OpenReview.net, 2020.
\newblock URL \url{https://openreview.net/forum?id=H1xFWgrFPS}.

\bibitem[Torralba et~al.(2008)Torralba, Fergus, and Freeman]{80mtiny}
Torralba, A., Fergus, R., and Freeman, W.~T.
\newblock 80 million tiny images: A large data set for nonparametric object and
  scene recognition.
\newblock \emph{IEEE Transactions on Pattern Analysis and Machine
  Intelligence}, 30\penalty0 (11):\penalty0 1958--1970, 2008.

\bibitem[Wang et~al.(2018)Wang, Zhu, Torralba, and
  Efros]{DBLP:journals/corr/abs-1811-10959}
Wang, T., Zhu, J., Torralba, A., and Efros, A.~A.
\newblock Dataset distillation.
\newblock \emph{CoRR}, abs/1811.10959, 2018.
\newblock URL \url{http://arxiv.org/abs/1811.10959}.

\bibitem[Xiao et~al.(2017)Xiao, Rasul, and Vollgraf]{xiao2017/online}
Xiao, H., Rasul, K., and Vollgraf, R.
\newblock Fashion-mnist: a novel image dataset for benchmarking machine
  learning algorithms, 2017.

\bibitem[Zagoruyko \& Komodakis(2016)Zagoruyko and
  Komodakis]{DBLP:conf/bmvc/ZagoruykoK16}
Zagoruyko, S. and Komodakis, N.
\newblock Wide residual networks.
\newblock In Wilson, R.~C., Hancock, E.~R., and Smith, W. A.~P. (eds.),
  \emph{Proceedings of the British Machine Vision Conference 2016, {BMVC} 2016,
  York, UK, September 19-22, 2016}. {BMVA} Press, 2016.
\newblock URL \url{http://www.bmva.org/bmvc/2016/papers/paper087/index.html}.

\bibitem[Zhao \& Bilen(2021)Zhao and Bilen]{DBLP:conf/icml/ZhaoB21}
Zhao, B. and Bilen, H.
\newblock Dataset condensation with differentiable siamese augmentation.
\newblock In Meila, M. and Zhang, T. (eds.), \emph{Proceedings of the 38th
  International Conference on Machine Learning, {ICML} 2021, 18-24 July 2021,
  Virtual Event}, volume 139 of \emph{Proceedings of Machine Learning
  Research}, pp.\  12674--12685. {PMLR}, 2021.
\newblock URL \url{http://proceedings.mlr.press/v139/zhao21a.html}.

\bibitem[Zhao et~al.(2021)Zhao, Mopuri, and Bilen]{DBLP:conf/iclr/ZhaoMB21}
Zhao, B., Mopuri, K.~R., and Bilen, H.
\newblock Dataset condensation with gradient matching.
\newblock In \emph{9th International Conference on Learning Representations,
  {ICLR} 2021, Virtual Event, Austria, May 3-7, 2021}. OpenReview.net, 2021.
\newblock URL \url{https://openreview.net/forum?id=mSAKhLYLSsl}.

\bibitem[Zintgraf et~al.(2017)Zintgraf, Cohen, Adel, and
  Welling]{DBLP:conf/iclr/ZintgrafCAW17}
Zintgraf, L.~M., Cohen, T.~S., Adel, T., and Welling, M.
\newblock Visualizing deep neural network decisions: Prediction difference
  analysis.
\newblock In \emph{5th International Conference on Learning Representations,
  {ICLR} 2017, Toulon, France, April 24-26, 2017, Conference Track
  Proceedings}. OpenReview.net, 2017.
\newblock URL \url{https://openreview.net/forum?id=BJ5UeU9xx}.

\end{thebibliography}
